\pdfoutput=1
\documentclass[11pt]{article}

\usepackage[final]{acl}

\usepackage{times}
\usepackage{latexsym}

\usepackage[T1]{fontenc}
\usepackage[utf8]{inputenc}

\usepackage{microtype}

\usepackage{inconsolata}

\usepackage{graphicx}

\usepackage{booktabs}
\usepackage{marvosym}
\usepackage{multirow}
\usepackage{colortbl}
\usepackage{array,xcolor}
\usepackage{colortbl}
\usepackage{lipsum}  % For filler text
\usepackage{placeins}

\definecolor{mygray}{gray}{0.9}

\usepackage{todonotes}
\title{Facilitating Long Context Understanding via \\Supervised Chain-of-Thought Reasoning}

\author{Jingyang Lin\textsuperscript{\rm 1,\rm 2{\bf†}}
        Andy Wong\textsuperscript{\rm 2},
        Tian Xia\textsuperscript{\rm 2}, 
        Shenghua He\textsuperscript{\rm 2}, 
        Hui Wei\textsuperscript{\rm 2,\rm 3{\bf†}}, 
        Mei Han\textsuperscript{\rm 2},
        Jiebo Luo\textsuperscript{\rm 1} \\
        \textsuperscript{\rm 1 }University of Rochester, NY, United States \\
        \textsuperscript{\rm 2 }PAII Inc., CA, United States \\
        \textsuperscript{\rm 3 }University of California, Merced, CA, United States\\
        \url{https://long-pai.github.io/}}

\definecolor{mygreen}{RGB}{160,250,160}
\definecolor{mydarkgreen}{RGB}{100,230,100}
\definecolor{myred}{RGB}{255,182,193}
\definecolor{mygrey}{RGB}{230,230,230}

\def\eg{\emph{e.g}.} 
\def\ie{\emph{i.e}.}

\makeatother

\input{def.set}

\begin{document}
\maketitle

\def\thefootnote{}\footnotetext{Corresponding author: \Letter~\href{mailto:jlin81@ur.rochester.edu}{jlin81@ur.rochester.edu}.}
\def\thefootnote{{\bf†}}\footnotetext{Work was done during the internship at PAII Inc.}\def\thefootnote{\arabic{footnote}}

\begin{abstract}
Recent advances in Large Language Models (LLMs) have enabled them to process increasingly longer sequences, ranging from 2K to 2M tokens and even beyond.
However, simply extending the input sequence length does not necessarily lead to effective long-context understanding.
In this study, we integrate Chain-of-Thought (CoT) reasoning into LLMs in a {\it supervised} manner to facilitate effective long-context understanding.
To achieve this, we introduce LongFinanceQA, a synthetic dataset in the financial domain designed to improve long-context reasoning.
Unlike existing long-context synthetic data, LongFinanceQA includes intermediate CoT reasoning before the final conclusion, which encourages LLMs to perform explicit reasoning, improving accuracy and interpretability in long-context understanding.
To generate synthetic CoT reasoning, we propose Property-based Agentic Inference (PAI), an agentic framework that simulates human-like reasoning steps, including property extraction,  retrieval, and summarization.
We evaluate PAI's reasoning capabilities by assessing GPT-4o-mini w/ PAI on the Loong benchmark, outperforming standard GPT-4o-mini by 20.0\%. Furthermore, we fine-tune LLaMA-3.1-8B-Instruct on LongFinanceQA, achieving a 28.0\% gain on Loong's financial subset.
\end{abstract}

\section{Introduction} \label{sec:intro}

\begin{figure}
    \centering
    \includegraphics[width=0.48\textwidth]{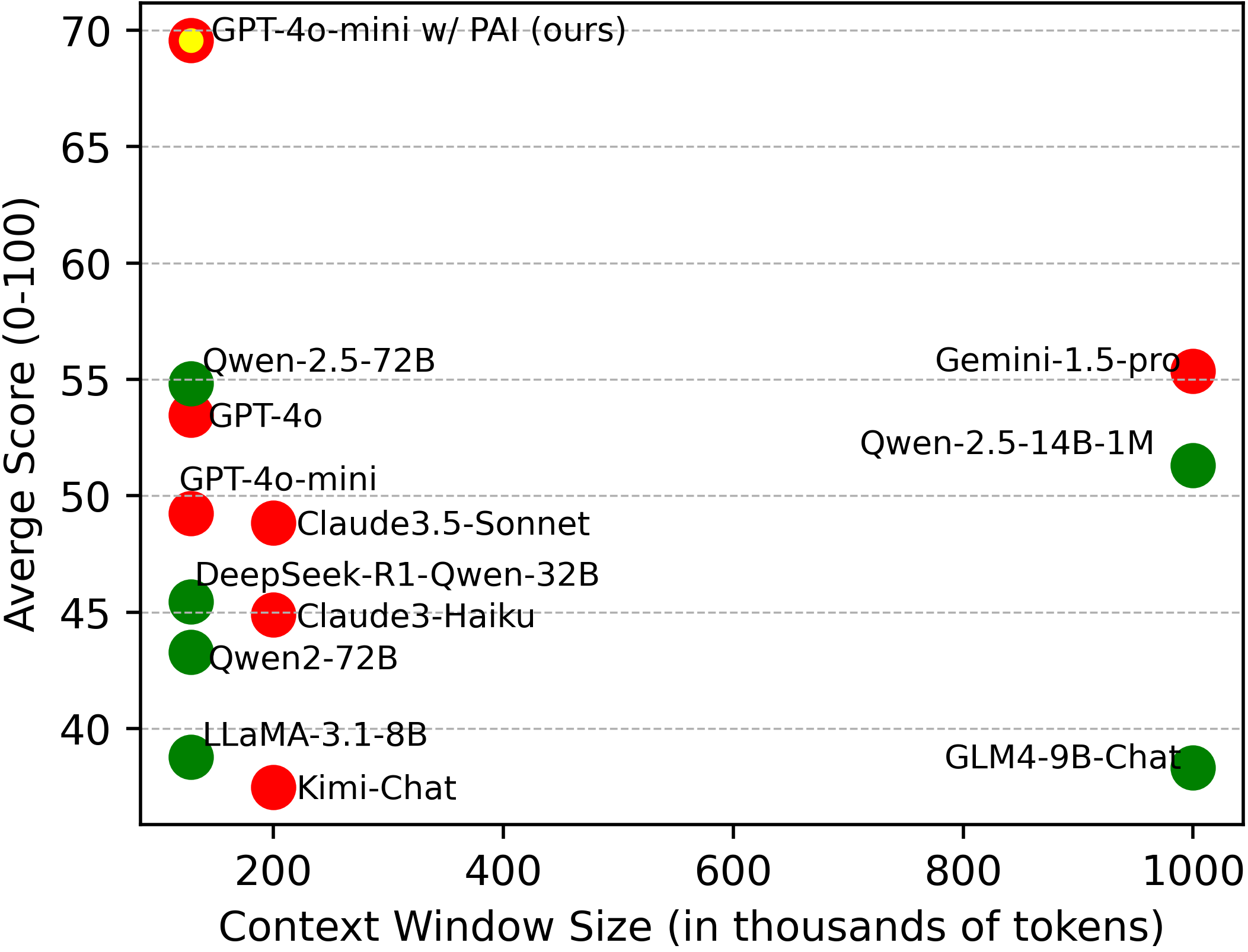}
    \caption{The hype of long-context large language models. The results shown are from the Loong benchmark, where green points refer to open-source LLMs and red points indicate closed-source LLMs. Our \textbf{GPT-4o-mini w/ PAI} stands out as the red open circle.}
    \label{fig:length_score}\vspace{-2mm}
\end{figure}

% bg: Long context understanding is challenging and important.
Long context understanding remains an evolving challenge~\cite{kocisky2018narrativeqa,wu2021recursively,bai2024longbench,wang2024leave} in natural language processing (NLP). 
Achieving long-context understanding requires processing long-form textual information, thereby enhancing a model’s ability to generate coherent, accurate, and contextually relevant responses~\cite{ibm2024llmcontext}.
Practical long context understanding has a potential impact on numerous applications, such as private document analysis~\cite{mukherjee2023feasibility}, large codebase understanding~\cite{nam2024using}, and multimodal content understanding~\cite{chandrasegaran2024hourvideo,lin2023videoxum,tang2023video,chen2023fine}.
Recent advancements in large language models~\cite{ouyang2022training,reid2024gemini,dubey2024llama} have significantly extended the input sequence length, ranging from 2K to 2M tokens, as shown in Figure~\ref{fig:trend} in Appendix \ref{sec:appendix}.

%% limitation
However, simply increasing the input sequence length does not necessarily improve the ability to comprehend the long content~\cite{yang2024qwen2,goldman2024really}. In particular, Figure~\ref{fig:length_score} presents the performance of various long-context LLMs on the Loong benchmark~\cite{wang2024leave}. The results suggest that, regardless of the maximum sequence length models can process, their performance remains similar, typically between 45\% and 55\%. This phenomenon exposes a hype that despite considerably enlarging context window size, the state-of-the-art LLMs still fail to perform satisfactorily in practical long-context problem-solving tasks.
%% Motivation
Therefore, instead of merely increasing the input sequence length, achieving effective long-context understanding remains an open challenge.

In parallel with advanced long-context architectures and techniques~\cite{peng2024yarn,liu2024ringattention,dao2024transformers}, constructing high-quality long-context training data remains essential yet underexplored.
Given the scarcity and high annotation costs of long-context data, generating high-quality synthetic data for long-context modeling is valuable and urgent. Early attempts~\cite{raffel2020exploring,fu2024data} simply pack all short-length data into long-context chunks without considering document boundaries.
Later works~\cite{zhang2024extending,he2024never} have introduced long-context QA tasks from both single- and multi-source perspectives.
Qwen-Agent~\cite{yang2024qwen2} develops an agentic system to further improve the quality of synthetic answers.

However, existing synthetic long-context data typically pair challenging questions with brief final answers for model training. This way overlooks a key difference between long-context and traditional QA tasks: practical long-context questions often require multi-step reasoning throughout the long content.
Without intermediate reasoning, LLMs struggle to learn effective patterns from the paired complex questions and brief answers.
We hypothesize that \emph{directly guiding models to generate brief answers without intermediate reasoning steps for long-context modeling will lead to suboptimal training}.
Instead, incorporating intermediate reasoning into synthetic data will help LLMs learn effective patterns and enhance training optimization.

% method
To validate this hypothesis, we introduce \emph{LongFinanceQA}, a novel long-context synthetic dataset constructed using financial data. Each sample in this dataset consists of a practical long-context question and the corresponding augmented answer along with intermediate chain-of-thought (CoT) reasoning steps.
In particular, we first collect 6,911 bilingual financial annual reports (\ie, English and Chinese) published before 2022. We then build a financial metric pool comprising key metrics commonly found in these reports, such as profit, cash flow, and debt. 
Based on these financial metrics, we generate 46,457 long-context questions that require single- or multi-source evidence. 
To generate reliable reasoning-augmented answers, we propose Property-based Agentic Inference (PAI), a comprehensive agentic framework. PAI leverages LLM-based agents to simulate human-like reasoning and operates in three steps:
1) \textbf{a property extraction agent} decomposes complex queries by identifying key properties, where each property consists of a metric and the corresponding subject;
2) \textbf{a property-based retrieval agent} retrieves relevant information from long documents for each identified property, and then generates intermediate findings by leveraging the property-based retrieved content;
3) \textbf{a summarization agent} synthesizes the intermediate findings into a coherent conclusion.
The synthetic reasoning results are formed by integrating outputs from the property extraction and retrieval stages, while the conclusion is derived from the summarization stage.
Finally, \emph{LongFinanceQA} comprises 46,457 long-context QA pairs with CoT reasoning over 6,911 financial reports.

Although PAI performs human-like reasoning in long-context scenarios, it relies on human-crafted design and multi-step inference. To simplify this inference process, we intend to transfer PAI's long-context reasoning ability to a large language model, LLaMA-3.1, via supervised fine-tuning on LongFinanceQA. The enhanced model, LongPAI, leverages CoT reasoning to handle long-context problems in a single step. This fine-tuning procedure is termed as \emph{Supervised CoT Reasoning}.

Empirically, we first evaluate the outcome of PAI on the Loong benchmark~\cite{wang2024leave}, involving challenging long-context tasks on three different domains. The results show that equipping GPT-4o-mini with PAI achieves a substantial 20\% improvement on the Loong as shown in Figure~\ref{fig:length_score}. Moreover, the effectiveness of PAI guarantees the quality of the synthetic data in LongFinanceQA. Meanwhile, the enhanced LLaMA-3.1 model, LongPAI, achieves a 28.0\% improvement on the \textit{Financial} subset of Loong.
Notably, in several scenarios, LongPAI even surpasses its teacher model PAI. This phenomenon emphasizes the importance of long-context modeling, contradicting recent arguments that \textit{the long-context problem can be solved by short language models}~\cite{qian2024long,chen2024long}.

Our main contributions are three-fold: 1) we introduce \emph{LongFinanceQA}, a long-context synthetic dataset for fine-tuning with 46,457 QA pairs featuring high-quality CoT reasoning from 6,911 bilingual financial annual reports; 2) to generate reasoning-augmented answers, we propose an agentic framework, \emph{Property-based Agentic Inference}, to mimic human behaviors; and 3) empirical results validate the effectiveness of PAI and supervised CoT reasoning, the quality of LongFinanceQA, and the importance of long-context modeling.

\section{Related Work}

\noindent \textbf{Long-context Synthetic Data}.
The scarcity of well-annotated long-context data makes high-quality synthetic data generation a valuable research direction. 
Early works~\cite{raffel2020exploring,fu2024data} concatenate short data into long-context fragments without considering the boundaries of the document. Large World Model~\cite{liu2024world} addresses this boundary issue with masked sequence packing to keep attention within documents. 
Subsequent studies~\cite{zhang2024extending,he2024never} construct single- and multi-source long-context QA pairs requiring evidence retrieval across multiple document positions.
Qwen2-Agent~\cite{yang2024qwen2} leverages multiple agents to enhance answer quality. Beyond conventional QA pairs, our study augments answers with intermediate reasoning steps, explicitly guiding language models in learning reasoning abilities for practical long-context scenarios.

\noindent \textbf{Long-context Large Language Models}. 
Two main approaches enhance long-context problem-solving: reduction-based and extension-based methods. Reduction-based methods compress input by preserving essential information, allowing LLMs to focus on relevant content. Techniques like Retrieval-Augmented Generation (RAG)~\cite{lewis2020retrieval} and task decomposition for book summarization\cite{wu2021recursively} follow this approach.
Extension-based methods expand the context window directly. RoPE~\cite{su2024roformer} introduces a foundational positional encoding widely used in long-context LLMs~\cite{peng2024yarn, yang2024qwen2, dubey2024llama}. Parallelism techniques~\cite{ren2021zero, liu2024ringattention} scale context capacity, enabling fully fine-tuning for long-context LLMs. This study leverages parallelism techniques to achieve long-context training.

%% Difference:
\noindent \textbf{Chain-of-Thought Resoning}. 
The CoT technique improves the reasoning abilities of language models by incorporating intermediate reasoning steps.
Early works~\cite{wei2022chain} present that prompting LLMs with step-by-step reasoning significantly improves performance on complex reasoning tasks. Following works~\cite{zelikman2022star,yao2024tree} explore structured CoT approaches, such as tree-based and self-consistent reasoning. This work incorporates CoT reasoning steps into fine-tuning data instead of well-crafted prompts. This approach enables the augmented data to explicitly supervise language models in reasoning skills.

\noindent \textbf{Agentic RAG}. Agentic RAG frameworks can be grouped into two paradigms~\cite{liang2025reasoning}: predefined reasoning and agentic reasoning RAG. Our proposed PAI framework is a hybrid of the two paradigms. Like predefined RAG systems~\cite{press-etal-2023-measuring,sarthi2024raptor}, PAI follows a structured, multi-step reasoning workflow. However, the components of PAI employ the techniques used in agentic reasoning RAG. Concretely, Property Extraction Agent and Proper-based Retrieval Agent use prompt-based approaches (\ie, function calling~\cite{openai_function_calling}) to generate structured outputs.
Agentic RAG systems~\cite{li-etal-2024-retrieval,yao2023react,openai_function_calling} typically treat follow-up queries as free-form text, which invites query drift, irrelevant results, and hard-to-audit reasoning. Different from agentic RAG, PAI instead converts each sub-query into a well-formed property, enabling verifiable retrieval, interpretable reasoning traces, and compact supervised training.
Beyond agentic RAG, this study further investigates transferring the capability of long-context reasoning into a lightweight language model via supervised CoT, resulting in the LongPAI model.

\section{Methodology}
In this section, we first introduce the problem formulation of the QA task enhanced by intermediate CoT reasoning in Section~\ref{sec:3.1}. Then, we present the procedure of LongFinanceQA dataset construction in Section~\ref{sec:3.2}. After that, we describe the fine-tuning details in Section~\ref{sec:3.3}.

\begin{figure*}
    \centering
    \includegraphics[width=1\textwidth]{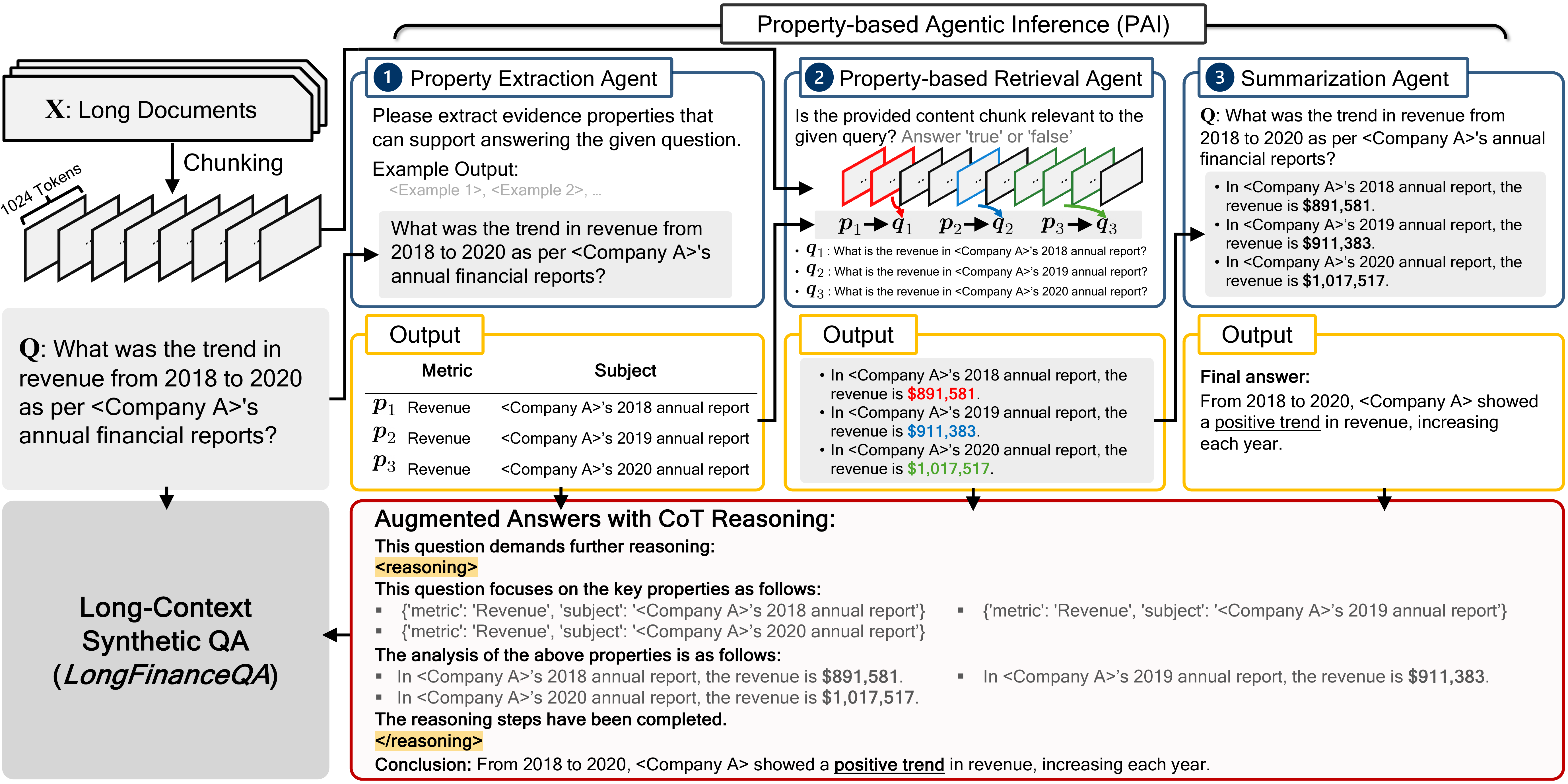}\vspace{-1mm}
    \caption{Overview of Property-based Agentic Inference (PAI), containing three stages. \textbf{A Property Extraction Agent} identifies key properties $\bp_i$ from the given query $\mathbf{Q}$, where each property consists of a measurable metric and its corresponding subject. Given the selected properties, \textbf{a Property-based Retrieval Agent} first transforms each property into a sub-query $\bq_i$ to retrieve relevant content chunks from long documents, yielding intermediate findings. \textbf{A Summarization Agent} integrates these intermediate findings to generate a comprehensive conclusion $\mathbf{A}$. After finishing PAI, we incorporate the output from the above three agents to produce reasoning-augmented answers. These augmented answers serve as the core contribution of the \textit{LongFinanceQA}.} \label{fig:pai}\vspace{-2mm}
\end{figure*}
\subsection{Problem Formulation} \label{sec:3.1}
Traditional QA tasks require language models to generate outputs $\mathbf{A}$ directly from a given query $\mathbf{Q}$ and the corresponding input content $\mathbf{X}$ by modeling the conditional probability:
\begin{equation}
    p_{\theta}(\mathbf{A}|\mathbf{X},\mathbf{Q}),
\end{equation}
where $\theta$ is the parameter of the language models.

Compared to traditional QA tasks, practical long-context QA tasks often require intermediate reasoning steps to analyze multiple pieces of evidence across long documents, and then derive the final answer. Thus, we formalize long-context QA tasks as a joint conditional probability of intermediate reasoning results $\mathbf{R}$ and the final answer $\mathbf{A}$:
\begin{equation}
    p_{\theta}(\mathbf{R}, \mathbf{A}|\mathbf{X},\mathbf{Q}).\label{eq:algo}
\end{equation}
As shown in Eq.(\ref{eq:algo}), this study aims to facilitate long-context understanding by guiding LLMs to first predict intermediate CoT reasoning steps before generating the final answer.
To achieve this, we construct a long-context synthetic dataset that explicitly supervises language models in learning intermediate reasoning. The data construction process can be factorized as follows:
\begin{equation}\small
    p_{\theta}(\mathbf{R}, \mathbf{A}|\mathbf{X},\mathbf{Q}) = p_{\theta}(\mathbf{R}|\mathbf{X},\mathbf{Q}) \cdot p_{\theta}(\mathbf{A}|\mathbf{X},\mathbf{Q},\mathbf{R}),\label{eq:data}
\end{equation}
where $p_{\theta}(\mathbf{R}|\mathbf{X},\mathbf{Q})$ indicates the generation of intermediate reasoning steps $\mathbf{R}$ given a query $\mathbf{Q}$ and input content $\mathbf{X}$, and then $p_{\theta}(\mathbf{A}|\mathbf{X},\mathbf{Q},\mathbf{R})$ is the process of incorporating the query and summarizing reasoning steps to produce the answer $\mathbf{A}$. The data construction follows the principle of Eq.(\ref{eq:data}).

\subsection{LongFinanceQA Dataset} \label{sec:3.2}\vspace{-1mm}
LongFinanceQA dataset is designed to generate practical long-context QA pairs with reasoning steps to effectively analyze long content. The finance domain was chosen for several reasons. First, annual financial reports are readily accessible and present complex long-context reasoning challenges. Moreover, finance is a data-driven field where accurate insights can drive critical decisions, making advancements in AI particularly valuable for real-world applications. We will introduce the data construction pipeline of LongFinanceQA as follows.

\noindent \textbf{Data Collection}. To begin, we collect bilingual financial annual reports (\ie, English and Chinese) dated before 2022 from open-source official websites, specifically the {SEC-10-K}\footnote{https://www.sec.gov/} and {cninfo}\footnote{http://www.cninfo.com.cn/} platforms. Documents from companies included in the Loong benchmark are then excluded. Next, we filter reports based on a token length range of 20K to 80K, determined using the GPT-4o tokenizer. Moreover, we prioritize companies with consistent annual reports over the years. In the end, 6,911 bilingual financial reports are selected.

\noindent \textbf{Diverse Query Generation}.
We first construct a financial metric pool containing key metrics commonly found in financial reports, such as profit, revenue, and cash flow. Then, we randomly pick several metrics in the metric pool and select a combination of financial reports. Given these financial metrics and the metadata from the selected documents (\eg, company name and year), we generate various long-context questions that require either single- or multi-source evidence. 
Next, we filter out questions whose corresponding combined documents exceed 256K tokens, as this surpasses the maximum token limit of our model. Finally, we obtain 46,457 practical long-context questions.
Please refer to Appendix~\ref{sec:appendix} for more details.

\noindent \textbf{Augmented Answer Generation with CoT Reasoning}.
Given long-context questions and their corresponding documents, our goal is to generate answers with CoT reasoning, following the principle of Eq. (\ref{eq:data}). In particular, we first generate step-by-step reasoning based on the question and documents, then integrate the reasoning steps into a conclusion.
Generally, long-context questions are challenging as they require models to retrieve multiple pieces of evidence scattered throughout long content and then integrate them for a global understanding~\cite{wang2024leave,edge2024local}. Inspired by this phenomenon, we intend to first extract key evidence points that support answering the question, then retrieve relevant information based on these points, and finally aggregate them into the conclusion. We term these supporting evidence points as ``\textit{Properties}''.
Following this methodology, we propose the \textbf{Property-based Agentic Inference (PAI)} framework, illustrated in Figure~\ref{fig:pai}, containing three steps: (1) property extraction, (2) property-based retrieval, and (3) summarization.

\textbf{Step 1: Property Extraction}. This step focuses on extracting a set of properties $\{\bp_i\}^{N_p}_{i=1}$ from the given query $\mathbf{Q}$, where $N_p$ denotes the number of properties. Each property consists of a metric (a measurable factor being analyzed) and its corresponding subject mentioned in the query $\mathbf{Q}$. For instance, given the query shown in Figure~\ref{fig:pai}, the metric is ``\textit{revenue}'' and subjects are ``<Company A>'s annual reports from different years (\eg, 2018, 2020, and 2022)'', which serve as the sources of information to determine the trend.

\textbf{Step 2: Property-based Retrieval}. After extracting the properties $\{\bp_i\}^{N_p}_{i=1}$ from the given query, this step aims to retrieve relevant information based on these properties. Specifically, each property is first transformed into a sub-query $\bq_i$, which is then matched against relevant content chunks. These chunks are derived from the original long documents $\mathbf{X}$, with each chunk limited to 1,024 tokens. Based on the sub-queries and their corresponding retrieved chunks, we can derive several intermediate findings (\ie, sub-answers).

\textbf{Step 3: Summarization}. In this final step, the original query $\mathbf{Q}$ and all intermediate findings from the second step are integrated to generate a comprehensive conclusion $\mathbf{A}$.

\begin{figure}
    \centering
    \includegraphics[width=0.43\textwidth]{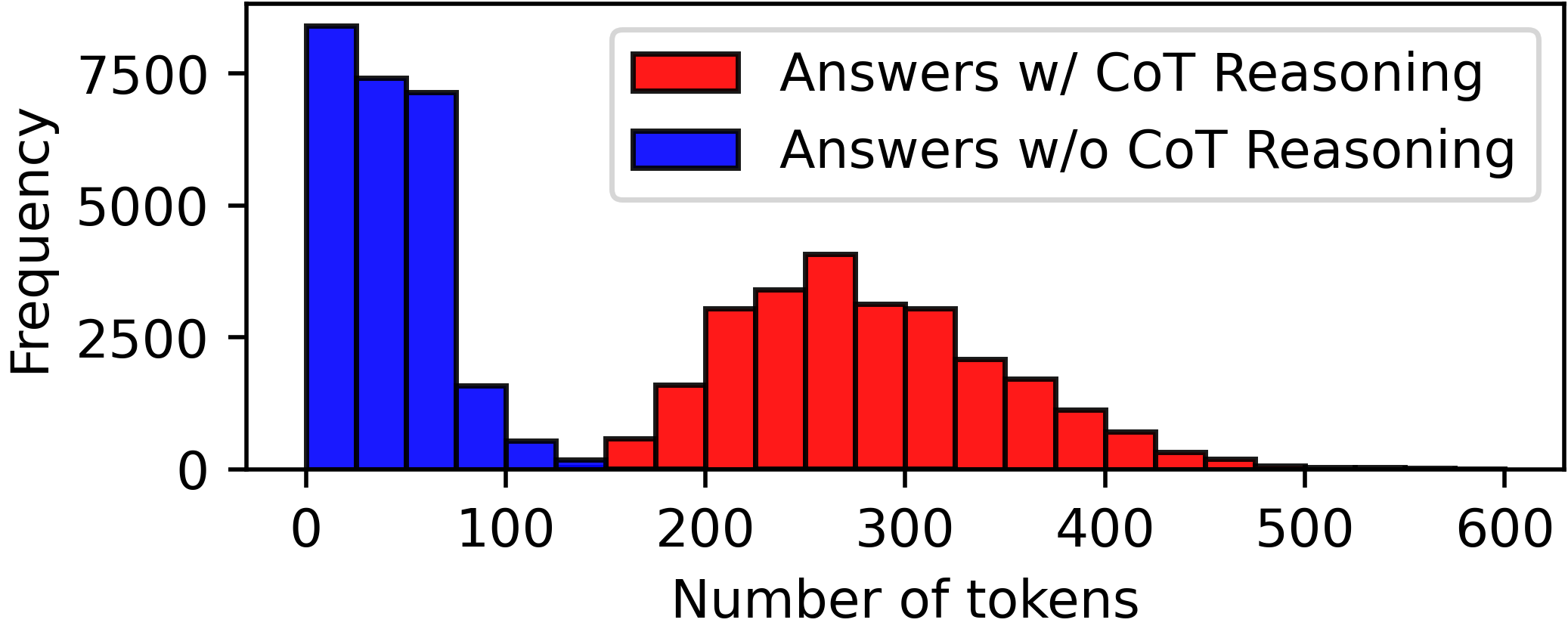} \vspace{-2mm}
    \caption{Token length distribution of answers with and without CoT reasoning from \textit{Multi-Source} QA pairs in the proposed LongFinanceQA dataset.} \label{fig:cot_stat} \vspace{-3mm}
\end{figure}
The PAI serves as a low-cost annotator for the LongFinanceQA dataset. Using PAI, we produce augmented answers with CoT reasoning by combining the intermediate reasoning results from the first two steps with the conclusion.
Figure~\ref{fig:cot_stat} illustrates the token length difference between answers with and without CoT reasoning, showing that reasoning-augmented answers are nearly 200 tokens longer on average.\footnote{The answers are tokenized by the LLaMA-3.1 tokenizer.} Please check out more data statistics in the Appendix~\ref{sec:appendix}.

\vspace{-2mm}
\subsection{Supervised CoT Reasoning} \label{sec:3.3}\vspace{-1mm}
Although PAI can behave like a human in long-context scenarios, it requires human-crafted design and multi-step inference.
To address this limitation, we seek to transfer the long-context reasoning capability of PAI to existing language models, enabling them to analyze long content in a single-step inference.
To this end, we fine-tune a language model on LongFinanceQA to explicitly guide it in learning CoT reasoning. This fine-tuning procedure is termed as \textit{Supervised CoT Reasoning}.
Specifically, we use LLaMA-3.1-8B-Instruct~\cite{dubey2024llama} as the base language model. Following~\cite{fu2024data}, we first extend the context window of LLaMA-3.1 from 128K to 262K through continued pretraining on 1.6B tokens, which consist of packed documents from Slimpajama~\cite{cerebras2023slimpajama}. After extending context length, we fine-tune the extended LLaMA-3.1 model (parameterized by $\theta$) on LongFinanceQA with reasoning-augmented answers.

Through this process, we obtain an enhanced LongPAI model by maximizing the log-likelihood of the predicted answer $\mathbf{Y}$ along with the CoT reasoning $\mathbf{R}$, conditioned on the given long documents $\mathbf{X}$ and the query $\mathbf{Q}$. 
The objective is calculated as:
\begin{equation}
\mathcal{L(\theta)} = \sum_{i=1}^{N_r+N_y}\log p_{\theta}(\mathcal{\mathbf{R}, \mathbf{Y} | \mathbf{X}, \mathbf{Q}}),
\end{equation}
where $N_r$ is the number of reasoning tokens and $N_y$ indicates the answering token, including those in the properties, sub-answers, and final answers. We mask non-answer positions during the fine-tuning, allowing for more efficient learning.
Although we only consider the LLaMA-3.1 model here, LongFinanceQA can also be used to fine-tune other language models (See Section~\ref{sec:4.1} for details). 
% Please refer to Section~\ref{sec:4.1} for more implementation details.

\section{Experiments}

\subsection{Experimental Setup}\label{sec:4.1}

\noindent \textbf{Evaluation Benchmarks}.
We evaluate long-context understanding using two practical benchmarks: Loong~\cite{wang2024leave} and $\infty$Bench~\cite{zhang2024bench}. Loong focuses on real-world multi-document question answering and comprises 1,600 test samples across four categories, including Spotlight Locating, Comparison, Clustering, and Chain of Reasoning. These tasks assess distinct capabilities in handling long-context tasks. $\infty$Bench facilitates multilingual evaluation, assessing models on English (En.QA) and Chinese (Zh.QA) question-answering tasks that require long-range dependency and reasoning beyond short-passage retrieval.
\begin{table*}[t]
\centering  
\caption{Data quality assessment for the long-context synthetic dataset (\textit{LongFinanceQA}) by measuring the performance of PAI on the Loong benchmark. \textit{AS} denotes \textit{Average Scores (0-100)}, and \textit{{PR}} represents the \textit{Perfect Rate} (0-1). \colorbox{mygreen}{Green} highlights the remarkable improvements over the base model (GPT-4o-mini).}\label{tab:pai_results}\vspace{-3mm}
\renewcommand{\arraystretch}{0.93}
\resizebox{\textwidth}{!}{
\begin{tabular}{llcccccccccc}
\toprule
\multirow{2}{*}{\textbf{Model}} & \multicolumn{1}{c}{\textbf{Context}} & \multicolumn{2}{c}{\textbf{Spotlight Locating}} & \multicolumn{2}{c}{\textbf{Comparison}} & \multicolumn{2}{c}{\textbf{Clustering}} & \multicolumn{2}{c}{\textbf{Chain of Reasoning}} & \multicolumn{2}{c}{\textbf{Overall}}\\ \cmidrule(r){3-4} \cmidrule(r){5-6} \cmidrule(r){7-8} \cmidrule(r){9-10} \cmidrule(r){11-12}
 & \multicolumn{1}{c}{\textbf{Length}} & \textbf{\textit{AS}} & \textbf{\textit{PR}} & \textbf{\textit{AS}} & \textbf{\textit{PR}} & \textbf{\textit{AS}} & \textbf{\textit{PR}} & \textbf{\textit{AS}} & \textbf{\textit{PR}} & \textbf{\textit{AS}} & \textbf{\textit{PR}} \\

\midrule
\multicolumn{12}{>{\columncolor[gray]{.88}}c}{\textit{Open-Source Long-Context Large Language Models}}  \\
LLaMA-3.1-8B-Instruct & 128K & 62.42 & 0.52 & 39.13 & 0.21 & 25.96 & 0.01 & 44.20 & 0.22 & 38.79 & 0.18 \\ 
DeepSeek-R1-Qwen-32B & 128K & 51.68 & 0.41 & 49.25 & 0.34 & 41.53 & 0.16 & 45.00 & 0.30 & 45.45 & 0.27 \\ 
Qwen2-72B-Instruct & 128K & 54.17 & 0.36 &42.38 & 0.20 & 36.71 & 0.04 & 47.76 & 0.18 & 43.29 & 0.15 \\ 
Qwen2.5-72B-Instruct & 128K & 65.08 & 0.55 & 51.90 & 0.30 & 46.07 & 0.08 & 64.43 & 0.40 & 54.83 & 0.28  \\ 
% LLaMA-3-8B-Instruct-262K & 262K & 48.77 & 0.30 & 29.33 & 0.11 & 19.93 & 0.01 & 21.27 & 0.07 & 27.52 & 0.10 \\ 
% GLM4-9B-Chat & 1000K & 57.35 & 0.47 & 40.38 & 0.20 & 28.52 & 0.02 & 39.94 & 0.16 & 38.31 & 0.16\\ 
Qwen2.5-14B-Instruct-1M & 1000K &  67.50 & 0.58 & 55.12 & 0.35 & 39.05 & 0.04 & 57.81 & 0.31 & 51.30 & 0.25  \\ 

\midrule
\multicolumn{12}{>{\columncolor[gray]{.88}}c}{\textit{Closed-Source Long-Context Large Language Models}}  \\

Kimi-Chat & 200K & 60.98 & 0.50 & 34.74 & 0.13 & 28.76 & 0.04 & 38.52 & 0.15 & 37.49 & 0.16 \\

% Claude3-Haiku & 200K & 68.68 & 0.59 & 42.10 & 0.21 & 35.04 & 0.02 & 47.59 & 0.17 & 44.88 & 0.19\\

Claude3.5-Sonnet & 200K & 58.45 & 0.49 & 54.21 & 0.35 & 45.77 & 0.07 & 43.92 & 0.25 & 48.85 & 0.23\\

GPT-4o & 128K & 73.95 &0.62 & 50.50 &0.28 &44.29 &0.09 &57.95 &0.28 &53.47 &0.26 \\

Gemini-1.5-pro & 1000K & 75.02 & 0.56 & 49.94 & 0.27 & 44.10 & 0.09 & 64.97 & 0.37 & 55.37 & 0.27 \\

\midrule
GPT-4o-mini (Base) & 128K & 59.46 & 0.49 & 51.90 & 0.27 & 34.55 & 0.04 & 64.28 & 0.39 & 49.25& 0.24 \\

GPT-4o-mini w/ PAI (\textit{ours}) & 128K & \cellcolor{mygreen}{\textbf{79.74}} & \cellcolor{mygreen}\textbf{0.67} & \cellcolor{mygreen}\textbf{67.60} & \cellcolor{mygreen}\textbf{0.46} & \cellcolor{mygreen}\textbf{62.80} & \cellcolor{mygreen}\textbf{0.27} & \cellcolor{mygreen}\textbf{75.46} & \cellcolor{mygreen}\textbf{0.57} & \cellcolor{mygreen}\textbf{69.58} & \cellcolor{mygreen}\textbf{0.44} \\
% \midrule
% LLaMA-3.1-8B-Instruct & 128K & 62.42 & 0.52 & 39.13 & 0.21 & 25.96 & 0.01 & 44.20 & 0.22 & 38.79 & 0.18\\ 
% \hspace{4mm} FT on LongFinance w/o Reasoning & 256K &  &  &  &  &  &  &  &  &  &  \\
% \hspace{4mm} FT on LongFinance w/ Reasoning & 256K &  &  &  &  &  &  &  &  &  &  \\
\bottomrule
\end{tabular}
}\vspace{-2mm}
\end{table*}

\noindent \textbf{Evaluation Metrics}. For evaluation, Loong employs GPT-4-Turbo as a judge, scoring model responses based on accuracy, hallucinations, and completeness on a scale of 0 to 100. Meanwhile, it introduces the Perfect Rate, measuring the proportion of responses achieving a perfect score. In $\infty$Bench, model performance is measured by the F1 score~\cite{zhang2024bench}. 

\noindent \textbf{Base Models}. 
We adopt GPT-4o-mini~\cite{achiam2023gpt} and LLaMA-3.1-8B-Instruct~\cite{dubey2024llama} as our base models. Specifically, GPT-4o-mini serves as the agent of the proposed Property-based Agentic Inference (PAI), while LLaMA-3.1-8B is used as the base model for LongPAI, which is fine-tuned on the LongFinanceQA.

\noindent \textbf{Implementation Details}.
To enable training on long sequences (> 250K), we employ several optimization techniques, including flash-attention-2~\cite{dao2023flashattention} and ring-attention~\cite{liu2024ringattention}. Furthermore, we adopt a zigzag sharding approach~\cite{zhu2024ringflash} within ring attention for more effective load distribution across multiple GPUs. This training setup allows us to fine-tune the large language models fully.
Using 8 A100 GPUs, the long-context training is completed in three days for fine-tuning.
In addition, in training the long-context LLMs, we adopt the rotary base scaling approach~\cite{liu2024scaling} and scale up the base value (\ie, rotary base of 1,247,820) to adapt RoPE to a longer context.
For optimization, we use a constant learning rate of 1e-5 for the entire training procedure.
Following the common practice~\cite{fu2024data}, we set the batch size to 16M tokens as mentioned in \cite{dubey2024llama}.
During inference, we set the temperature as zero to eliminate the randomness. We also increase the maximum output tokens to 1,024 since the CoT reasoning requires more output tokens.

\subsection{Main Results}
In this section, we first assess the quality of our long-context synthetic data (LongFinanceQA) by evaluating the performance of the proposed PAI on the Loong benchmark. Next, we compare the enhanced long-context language model (LongPAI) with its base LLaMA-3.1 model and other state-of-the-art LLMs on the \textit{Finance} subset of Loong.

\noindent \textbf{Data Quality Assessment}.
Reasoning-augmented answers in LongFinanceQA are automatically generated by the PAI framework. To assess the quality of these synthetic answers, we evaluate the annotator, PAI, on the Long benchmark, using GPT-4o-mini as the agent within the PAI, referred to as GPT-4o-mini w/ PAI.
Table~\ref{tab:pai_results} shows that the GPT-4o-mini-based PAI framework significantly enhances the base model’s performance, demonstrating the effectiveness of the PAI.
In particular, compared to the standard GPT-4o-mini model, the overall average score improves by 20.3\%, with substantial gains in key tasks such as spotlight locating (+20.2\%), comparison (+15.7\%), clustering (+29.2\%), and chain of reasoning (+10.2\%).
Although the basic GPT-4o-mini is not the strongest model, GPT-4o-mini w/ PAI outperforms the state-of-the-art closed-source model, Geneni-1.5-pro~\cite{reid2024gemini}, by over 15\%. 
Moreover, the superior performance in the Spotlight Locating task (\ie, single-source QA task) \textit{highlights the quality of intermediate reasoning results generated by the PAI}, as predictions in this task serve as essential reasoning steps for the other three multi-source QA tasks.
Consequently, the strong performance on the Loong demonstrates \textit{the capability of PAI as a reliable annotator, ensuring high-quality synthetic data in long-context scenarios}.

\noindent \textbf{Human Evaluation on LongFinanceQA}. To further quantify the data quality, we manually annotated 200 randomly selected LongFinanceQA questions, including both single-source and multi-source questions. Then, 10 volunteers score each response on a 1-to-5 scale (1 = completely incorrect, 5 = completely correct). Furthermore, volunteers rate the correctness of intermediate reasoning steps on a 1-to-5 scale for each multi-source question.  The averaged scores above 4.0 can be interpreted as largely correct. Table~\ref{tab:human_eval} shows that both final answers and their supporting reasoning trajectories are consistently reliable. It provides concrete evidence that the performance improvements primarily arise from high-quality synthesized data, rather than from patterns in noisy samples.

\noindent \textbf{Method Comparisons on Existing Benchmarks}.
To evaluate the effectiveness of supervised CoT reasoning 
on long-context modeling, we measure the performance of the enhanced LongPAI model on two well-known long-context understanding benchmarks, including Loong and $\infty$Bench. 
First, we evaluate the LongPAI on an in-domain benchmark, namely the \textit{Finance} subset of Loong.
Table~\ref{tab:longpai_results} presents that supervised CoT reasoning enables LongPAI to outperform its base model, LLaMA-3.1-8B-Instruct, by 28.0\% in overall results. Furthermore, LongPAI exhibits a 30\% improvement on subsets with longer content (\ie, 200K-250K). Also, LongPAI achieves a competitive performance against existing state-of-the-art language models. Remarkably, LongPAI is comparable to its teacher model, GPT-4o-mini w/ PAI (73.94\% vs. 75.56\%). In some settings, LongPAI even surpasses its teacher model. This finding highlights the significance of long-context modeling. Meanwhile, this finding strongly challenges the recent claim that \textit{the long-context problem can be adequately addressed by short language models}~\cite{qian2024long,chen2024long}. In other words, certain long-context problems require long-context modeling, as short language models struggle to analyze and reason effectively over reduced or retrieved information. At the same time, we evaluate the LongPAI on $\infty$Bench. The results in Table~\ref{tab:infbench} show that the LongPAI outperforms its base model even on the out-of-domain benchmark.

\begin{table}[t!]
\centering
\caption{Human evaluation on LongFinanceQA dataset.} \label{tab:human_eval}\vspace{-3mm}
\renewcommand{\arraystretch}{0.9}
\resizebox{0.48\textwidth}{!}{
\small
\begin{tabular}{lc}
\toprule
\textbf{Subset} & \textbf{Score} \\
\midrule
Single-source QA & 4.45 $\pm$ 0.21 \\
Multi-source QA & 4.24 $\pm$ 0.23\\
Intermediate Reasoning Trajectories & 4.30 $\pm$ 0.22 \\
\midrule
Overall & 4.29 $\pm$ 0.19\\
\bottomrule
\end{tabular}
}\vspace{-3mm}
\end{table}

\begin{table*}[t]
\centering
\caption{Performance on \textit{Financial} subset of Loong benchmark. \textit{AS} represents \textit{Avg Scores (0\textasciitilde100)} and \textit{{PR}} denotes \textit{Perfect Rate} (0\textasciitilde1). \colorbox{mygreen}{Green} highlights improvements over the base model. {Full results are shown in Appendix~\ref{sec:appendix}}.}\vspace{-2.5mm}
\renewcommand{\arraystretch}{0.93}
\resizebox{\textwidth}{!}{
\begin{tabular}{llcccccccccc}
\toprule
\multirow{2}{*}{\textbf{Model}} & \multicolumn{1}{c}{\textbf{Context}} & \multicolumn{2}{c}{\textbf{Spotlight Locating}} & \multicolumn{2}{c}{\textbf{Comparison}} & \multicolumn{2}{c}{\textbf{Clustering}} & \multicolumn{2}{c}{\textbf{Chain of Reasoning}} & \multicolumn{2}{c}{\textbf{Overall}}\\ \cmidrule(r){3-4} \cmidrule(r){5-6} \cmidrule(r){7-8} \cmidrule(r){9-10} \cmidrule(r){11-12}
 & \multicolumn{1}{c}{\textbf{Length}} & \textbf{\textit{AS}} & \textbf{\textit{PR}} & \textbf{\textit{AS}} & \textbf{\textit{PR}} & \textbf{\textit{AS}} & \textbf{\textit{PR}} & \textbf{\textit{AS}} & \textbf{\textit{PR}} & \textbf{\textit{AS}} & \textbf{\textit{PR}} \\

\midrule
\multicolumn{12}{>{\columncolor[gray]{.88}}c}{\textbf{$\mathtt{All\ Set}$ (10K-250K)}}  \\
% DeepSeek-R1-Qwen-32B & 128K &  53.66 & 0.46 & 52.19 & 0.39 & 39.76 & 0.17 & 65.15 & 0.51 & 49.92 & 0.34 \\
Qwen2-72B-Instruct & 128K &   59.80 & 0.47 & 61.12 & 0.43 & 34.32 & 0.06 & 74.68 & 0.50 & 53.20 & 0.32  \\
Qwen2.5-72B-Instruct & 128K &  71.07 & 0.63 & 59.14 & 0.41 & 38.23 & 0.08 & 81.09 & 0.60 & 57.36 & 0.36  \\
% LLaMA-3-8B-Instruct-262K & 262K  &  58.60 & 0.41 & 33.12 & 0.16 & 20.04 & 0.01 & 35.10 & 0.09 & 34.41 & 0.15 \\
% GLM4-9B-Chat & 1000K & 72.69 & 0.60 & 49.31 & 0.32 & 23.41 & 0.02 & 60.77 & 0.28 & 46.71 & 0.27 \\
Qwen-2.5-14B-Instruct-1M & 1000K & 78.83 & 0.72 & 65.27 & 0.50 & 36.24 & 0.07 & 79.10 & 0.64 & 59.78 & 0.41 \\
GPT-4o & 128K &  88.23 & 0.84 & 62.90 & 0.48 & 45.51 & 0.17 & 69.40 & 0.43 & 63.05 & 0.44 \\
\midrule
GPT-4o-mini (Base) & 128K & 70.90 & 0.59 & 59.37 & 0.38 & 36.33 & 0.06 & 79.58 & 0.58 & 56.50 & 0.34 \\
GPT-4o-mini w/ PAI (\textit{ours}) & 128K & \cellcolor{mygreen}91.07 & \cellcolor{mygreen}0.83 & \cellcolor{mygreen}74.40 & \cellcolor{mygreen}0.58 & \cellcolor{mygreen}61.55 & \cellcolor{mygreen}0.32 & \cellcolor{mygreen}89.63 & \cellcolor{mygreen}0.78 & \cellcolor{mygreen}75.56 & \cellcolor{mygreen}0.57 \\
\midrule
LLaMA-3.1-8B-Instruct (Base) & 128K & 67.84 & 0.56 & 47.12 & 0.30 & 24.62 & 0.02 & 63.63 & 0.34 & 45.88 & 0.26 \\ 
LongPAI (\textit{ours}) & 262K & \cellcolor{mygreen}89.79 & \cellcolor{mygreen}0.84 & \cellcolor{mygreen}71.69 & \cellcolor{mygreen}0.60 & \cellcolor{mygreen}59.71 & \cellcolor{mygreen}0.32 & \cellcolor{mygreen}90.28 & \cellcolor{mygreen}0.83 & \cellcolor{mygreen}73.94 & \cellcolor{mygreen}0.58 \\

\midrule
\multicolumn{12}{>{\columncolor[gray]{.88}}c}{\textbf{$\mathtt{Set3}$ (100K-200K)}}  \\
% DeepSeek-R1-Qwen-32B & 128K &  41.73 & 0.33 & 39.56 & 0.23 & 25.67 & 0.06 & 55.14 & 0.37 & 37.35 & 0.21 \\
Qwen2-72B-Instruct & 128K &  47.00 & 0.33 & 48.07 & 0.27 & 25.79 & 0.00 & 69.37 & 0.34 & 42.98 & 0.20  \\
Qwen2.5-72B-Instruct & 128K &  60.47 & 0.48 & 49.00 & 0.28 & 30.61 & 0.01 & 76.54 & 0.46 & 48.99 & 0.26  \\
% LLaMA-3-8B-Instruct-262K & 262K  &  54.14 & 0.40 & 22.93 & 0.05 & 15.43 & 0.00 & 29.00 & 0.00 & 28.58 & 0.11 \\
% GLM4-9B-Chat & 1000K & 74.75 & 0.65 & 41.63 & 0.24 & 21.99 & 0.01 & 49.86 & 0.17 & 43.58 & 0.25 \\
Qwen-2.5-14B-Instruct-1M & 1000K &  74.33 & 0.68 & 54.64 & 0.35 & 30.72 & 0.01 & 73.71 & 0.51 & 53.47 & 0.33 \\
GPT-4o & 128K &  87.25 & 0.83 & 46.00 & 0.31 & 36.68 & 0.08 & 64.57 & 0.40 & 54.79 & 0.36 \\
\midrule
GPT-4o-mini (Base) & 128K & 63.05 & 0.53 & 53.48 & 0.24 & 29.80 & 0.01 & 72.37 & 0.46 & 50.03 & 0.26 \\
GPT-4o-mini w/ PAI (\textit{ours}) & 128K & \cellcolor{mygreen}94.08 & \cellcolor{mygreen}0.83 & \cellcolor{mygreen}74.13 & \cellcolor{mygreen}0.63 & \cellcolor{mygreen}55.78 & \cellcolor{mygreen}0.20 & \cellcolor{mygreen}87.71 & \cellcolor{mygreen}0.77 & \cellcolor{mygreen}74.21 & \cellcolor{mygreen}0.55\\
\midrule
LLaMA-3.1-8B-Instruct (Base) & 128K & 63.38 & 0.47 & 36.04 & 0.19 & 20.28 & 0.00 & 62.49 & 0.26 & 40.45 & 0.20 \\ 
LongPAI (\textit{ours}) & 262K & \cellcolor{mygreen}94.17 & \cellcolor{mygreen}0.90 & \cellcolor{mygreen}63.76 & \cellcolor{mygreen}0.49 & \cellcolor{mygreen}51.83 & \cellcolor{mygreen}0.24 & \cellcolor{mygreen}88.66 & \cellcolor{mygreen}0.77 & \cellcolor{mygreen}70.00 & \cellcolor{mygreen}0.54 \\

\midrule
\multicolumn{12}{>{\columncolor[gray]{.88}}c}{\textbf{$\mathtt{Set4}$ (200K-250K)}}  \\
% DeepSeek-R1-Qwen-32B & 128K &  18.33 & 0.11 & 10.25 & 0.05 & 11.27 & 0.00 & 26.00 & 0.07 & 15.52 & 0.05 \\
Qwen2-72B-Instruct & 128K &  41.85 & 0.19 & 39.75 & 0.15 & 29.17 & 0.03 & 41.67 & 0.07 & 37.23 & 0.11  \\
Qwen2.5-72B-Instruct & 128K &  57.48 & 0.44 & 49.50 & 0.30 & 27.33 & 0.00 & 55.00 & 0.13 & 45.51 & 0.22  \\
% LLaMA-3-8B-Instruct-262K & 262K  &  34.19 & 0.07 & 20.00 & 0.05 & 20.31 & 0.00 & 20.71 & 0.00 & 24.61 & 0.03 \\
% GLM4-9B-Chat & 1000K & 40.85 & 0.19 & 29.50 & 0.05 & 18.13 & 0.00 & 25.73 & 0.00 & 28.51 & 0.07 \\
Qwen-2.5-14B-Instruct-1M & 1000K &  59.44 & 0.41 & 37.00 & 0.20 & 27.33 & 0.00 & 31.67 & 0.00 & 39.57 & 0.16 \\
GPT-4o & 128K &  69.26 & 0.56 & 50.50 & 0.35 & 30.70 & 0.00 & 50.67 & 0.07 & 49.58 & 0.25 \\
\midrule
GPT-4o-mini (Base) & 128K & 48.37 & 0.26 & 50.00 & 0.30 & 28.70 & 0.00 & 48.33 & 0.07 & 42.30 & 0.15 \\
GPT-4o-mini w/ PAI (\textit{ours}) & 128K & \cellcolor{mygreen}82.78 & \cellcolor{mygreen}0.70 & \cellcolor{mygreen}63.50 & \cellcolor{mygreen}0.35 & \cellcolor{mygreen}48.00 & \cellcolor{mygreen}0.17 & \cellcolor{mygreen}76.00 & \cellcolor{mygreen}0.53 & \cellcolor{mygreen}66.14 & \cellcolor{mygreen}0.42\\
\midrule
LLaMA-3.1-8B-Instruct (Base) & 128K & 40.74 & 0.30 & 35.85 & 0.20 & 19.77 & 0.00 & 28.73 & 0.00 & 30.88 & 0.13 \\ 
LongPAI (\textit{ours})  & 262K & \cellcolor{mygreen}71.48 & \cellcolor{mygreen}0.59 & \cellcolor{mygreen}56.50 & \cellcolor{mygreen}0.45 & \cellcolor{mygreen}46.83 & \cellcolor{mygreen}0.17 & \cellcolor{mygreen}76.00 & \cellcolor{mygreen}0.67 & \cellcolor{mygreen}60.92 & \cellcolor{mygreen}0.43  \\ 
\bottomrule
\end{tabular}
}
\label{tab:longpai_results}\vspace{-1mm}
\end{table*}
\begin{table}[t!]
\centering
\caption{Comparison on En.QA and Zh.QA of $\infty$Bench. $^*$ indicates results borrowed from~\cite{zhang2024bench}.}\label{tab:infbench}\vspace{-3mm}
\renewcommand{\arraystretch}{0.9}
\resizebox{0.48\textwidth}{!}{
\small
\begin{tabular}{lcc}
\toprule
\textbf{} & \textbf{En.QA} & \textbf{Zh.QA} \\
\midrule
YaRN-Mistra$^*$ & 9.55 & 16.98 \\
Kimi-Chat$^*$ & 16.52 & 18.62 \\
Claude 2$^*$ & 11.97 & 10.53 \\
GPT-4$^*$ & 22.22 & 23.06 \\
\midrule
LLaMA-3.1-8B-Instruct (Base)& 27.11 & 29.77 \\
LongPAI (ours)& \textbf{32.65} & \textbf{32.88} \\
\bottomrule
\end{tabular}
}\vspace{-3mm}
\end{table}

\subsection{Discussion}
\noindent \textbf{Ablation Study on Supervised CoT Reasoning}. 
To further analyze the impact of supervised CoT reasoning, we fine-tune the base LLaMA-3.1 model on LongFinanceQA while excluding CoT reasoning steps from the augmented answers, resulting in a new model, LongPAI$^{\S}$.
Unlike the original LongPAI, LongPAI$^{\S}$ directly predicts the final answer, skipping intermediate reasoning steps.
Table~\ref{tab:cot_ablation} shows that LongPAI significantly outperforms LongPAI$^{\S}$ over different input lengths. This result strongly supports the hypothesis mentioned in Section~\ref{sec:intro}, which argues that \emph{directly guiding models to generate brief answers without intermediate reasoning steps for long-context modeling will lead to suboptimal training} (\textbf{finding 1}).
Moreover, Table~\ref{tab:cot_ablation} presents several interesting findings.
First, while LongPAI$^{\S}$ performs comparably to LongPAI on short content (10K–50K tokens), its performance declines significantly on longer content, which means \emph{CoT reasoning is necessary for long-context modeling} (\textbf{finding 2}). Furthermore, LongPAI$^{\S}$ performs well in single-source questions (Spotlight Locating) but struggles with multi-source questions (Comparison, Clustering, and Chain of Reasoning) as input length increases, demonstrating that \emph{CoT reasoning benefits complex long-context problem-solving} (\textbf{finding 3}). In sum, all these findings reaffirm the importance of the reasoning capability for long-context modeling.

\noindent \textbf{Comparison of Various Inference Frameworks}. We compare PAI with four relevant inference frameworks: PAI$^-$, Self-Route~\cite{li-etal-2024-retrieval}, Self-Ask~\cite{press-etal-2023-measuring}, and RAG~\cite{lewis2020retrieval}.
PAI$^-$ is a variant of PAI that generates sub-questions directly, rather than first extracting properties and then generating sub-questions.
Table~\ref{tab:inference_comparison} presents that PAI$^-$ generally outperforms the base GPT-4o-mini, except on $\mathtt{Set 1}$. However, there is a gap between PAI$^-$ and PAI, highlighting the superiority of the Property Extraction Agent.
RAG follows a two-step process: it first retrieves the top $K$ chunks relevant to a given query, and then uses the retrieved chunks to generate an answer. Following Loong~\cite{wang2024leave}, we adopt the \textit{BGE}~\cite{chen-etal-2024-m3} as the embedding choice and set $K$ to 50, selecting from 5, 10, 30, and 50. Table~\ref{tab:inference_comparison} shows that RAG struggles with long-context problems, a conclusion also reached by previous work~\cite{wang2024leave}.
Furthermore, we compare the proposed PAI framework with both predefined reasoning RAG (Self-Route) and agentic reasoning RAG (Self-Ask). The results demonstrate that PAI consistently outperforms the other two agentic RAG methods, especially on longer contexts (Set3 and Set4), highlighting a significant superiority of the PAI framework under the scenario of long-context understanding.

\begin{table}[t!]
\centering
\caption{Ablation study on  \textit{Supervised CoT Reasoning}. Symbol $^{\S}$ represents LongPAI without supervised CoT reasoning during fine-tuning. Average Scores (0-100) are evaluated by GPT-4-Turbo. \colorbox{mygreen}{Green} highlights the remarkable improvements over the base LLaMA-3.1-8B, while \colorbox{myred}{Red} indicates a decline.
Abbreviations: \textbf{S.L.} (Spotlight Locating), \textbf{Comp.} (Comparison), \textbf{Clust.} (Clustering), and \textbf{Chain.} (Chain of Reasoning). } 
\label{tab:cot_ablation}\vspace{-3mm}
\renewcommand{\arraystretch}{1}
\resizebox{0.49\textwidth}{!}{
\begin{tabular}{lccccc}
\toprule
\textbf{Method} & \textbf{S.L.} & \textbf{Comp.} & \textbf{Clust.} & \textbf{Chain.} & \textbf{Overall} \\
\midrule

\multicolumn{6}{>{\columncolor[gray]{.88}}c}{\textbf{$\mathtt{Set1}$ (10K-50K)}}  \\
LLaMA-3.1-8B & 89.13 & 72.33 & 31.77 & 74.0 & 60.50 \\
LongPAI$^{\S}$ & \cellcolor{mygreen}93.39 & \cellcolor{myred}68.67 & \cellcolor{mydarkgreen}79.38 & \cellcolor{mygreen}83.80 & \cellcolor{mygreen}79.82 \\
LongPAI & \cellcolor{mydarkgreen}97.30 & \cellcolor{mydarkgreen}90.17 & \cellcolor{mygreen}72.88 & \cellcolor{mydarkgreen}94.00 & \cellcolor{mydarkgreen}85.42\\
\midrule
\multicolumn{6}{>{\columncolor[gray]{.88}}c}{\textbf{$\mathtt{Set2}$ (50-100K)}}  \\
LLaMA-3.1-8B & 80.58 & 51.11 & 27.39 & 75.12 & 51.13 \\
LongPAI$^{\S}$ & \cellcolor{mygreen}82.12 & \cellcolor{myred}40.40 & \cellcolor{myred}11.24 & \cellcolor{myred}50.25 & \cellcolor{myred}38.11 \\
LongPAI & \cellcolor{mydarkgreen}91.25 & \cellcolor{mydarkgreen}76.27 & \cellcolor{mydarkgreen}66.02 & \cellcolor{mydarkgreen}96.12 & \cellcolor{mydarkgreen}78.19 \\
\midrule
\multicolumn{6}{>{\columncolor[gray]{.88}}c}{\textbf{$\mathtt{Set3}$ (100K-200K)}}  \\
LLaMA-3.1-8B & 63.38 & 36.04 & 20.28 & 62.49 & 40.45 \\
LongPAI$^{\S}$ & \cellcolor{myred}59.58 & \cellcolor{myred}30.43 & \cellcolor{myred}5.47 & \cellcolor{myred}37.43 & \cellcolor{myred}29.46 \\
LongPAI & \cellcolor{mydarkgreen}94.17 & \cellcolor{mydarkgreen}63.76 & \cellcolor{mydarkgreen}51.83 & \cellcolor{mydarkgreen}88.66 & \cellcolor{mydarkgreen}70.00 \\
\midrule
\multicolumn{6}{>{\columncolor[gray]{.88}}c}{\textbf{$\mathtt{Set4}$ (200K-250K)}}  \\
LLaMA-3.1-8B & 40.74 & 35.85 & 19.77 & 28.73 & 30.88 \\
LongPAI$^{\S}$ & \cellcolor{mygreen}43.37 & \cellcolor{myred}22.00 & \cellcolor{myred}5.60 & \cellcolor{myred}17.20 & \cellcolor{myred}22.14 \\
LongPAI & \cellcolor{mydarkgreen}71.48 & \cellcolor{mydarkgreen}56.50 & \cellcolor{mydarkgreen}46.83 & \cellcolor{mydarkgreen}76.00 & \cellcolor{mydarkgreen}60.92 \\
\bottomrule
\end{tabular}
}\vspace{-2mm}
\end{table}

\noindent \textbf{Efficiency Analysis on PAI and LongPAI}. The PAI framework relies on multi-step inference, whereas LongPAI achieves results with a single inference step. We adopt the number of input tokens processed on the Loong benchmark (\ie, 1,600 samples) as a metric to compare the efficiency between PAI and LongPAI. In this comparison, \emph{PAI processes 3.53B tokens in total, determined by the GPT-4o tokenizer, whereas LongPAI requires only 112M tokens, \textbf{less than 3\% of PAI's total}}. Despite PAI delivering the highest overall performance, LongPAI stands out as the far more efficient approach, demonstrating a dramatic reduction in computational cost.

\section{Conclusion}

In this work, we introduce LongFinanceQA, a novel long-context synthetic dataset featuring reasoning-augmented answers for practical long-context questions. To generate these enriched answers, we develop the Property-based Agentic Inference (PAI) framework, which simulates human-like reasoning through property extraction, retrieval, and summarization. Empirical analysis demonstrates the effectiveness of PAI and ensures the high quality of the LongFinanceQA dataset. Beyond PAI, we explicitly guide a lightweight language model in learning CoT reasoning via fine-tuning on LongFinanceQA. A series of empirical results underscores the importance of reasoning-augmented long-context modeling.

\begin{table}[t!]
\centering
\caption{Comparison of inference frameworks on the Loong Benchmark. Performance is evaluated by GPT-4-Turbo across four sets with different context sizes: \textbf{$\mathtt{Set 1}$} (10–50K), \textbf{$\mathtt{Set 2}$} (50–100K), \textbf{$\mathtt{Set 3}$} (100–200K), and \textbf{$\mathtt{Set 4}$} (200–250K). \colorbox{mygreen}{Green} indicates an improvement over base GPT-4o-mini. \colorbox{myred}{Red} denotes a decline.} 
\label{tab:inference_comparison}\vspace{-3mm}
\renewcommand{\arraystretch}{1}
\resizebox{0.48\textwidth}{!}{
\begin{tabular}{lccccc}
\toprule
\textbf{Method} & {\textbf{$\mathtt{Set 1}$}} & {\textbf{$\mathtt{Set 2}$}} & {\textbf{$\mathtt{Set 3}$}} & {\textbf{$\mathtt{Set 4}$}} & \textbf{Overall} \\
\midrule
GPT-4o-mini & 65.05 & 50.63 & 45.41 & 31.61 & 49.25 \\
\hspace{8mm} w/ PAI & \cellcolor{mydarkgreen}78.69 & \cellcolor{mydarkgreen}69.62 & \cellcolor{mydarkgreen}68.95 & \cellcolor{mydarkgreen}58.12 & \cellcolor{mydarkgreen}69.58 \\
\hspace{8mm} w/ PAI$^-$ & \cellcolor{myred}60.52 & \cellcolor{mygreen}57.69 & \cellcolor{mygreen}54.89 & \cellcolor{mygreen}37.99 & \cellcolor{mygreen}54.56 \\
\hspace{8mm} w/ Self-Ask & \cellcolor{myred}62.44 & \cellcolor{mygreen}52.37 & \cellcolor{mygreen}46.91 & \cellcolor{mygreen}34.53 & \cellcolor{mygreen}50.17 \\
\hspace{8mm} w/ Self-Route & \cellcolor{mygreen}68.79 & \cellcolor{mygreen}55.24 & \cellcolor{mygreen}49.16 & \cellcolor{mygreen}34.44 & \cellcolor{mygreen}53.13 \\
\hspace{8mm} w/ RAG & \cellcolor{myred}64.67 & \cellcolor{myred}41.21 & \cellcolor{myred}31.80 & \cellcolor{myred}22.08 & \cellcolor{myred}40.35 \\
\bottomrule
\end{tabular}
}\vspace{-3mm}
\end{table}

\section*{Limitations}\vspace{-2mm}
In this study, we investigate the effectiveness of supervised CoT reasoning using reasoning-augmented long-context synthetic data (LongFinanceQA). While effective, our approach has certain limitations.
First, although the reasoning-enhanced LongPAI model demonstrates significant improvement in the financial domain, its ability to generalize to broader long-context scenarios remains uncertain since overfitting is an inherent issue that stems from supervised fine-tuning~\cite{li2025preserving}. To address this, our future work will explore the impact of diverse data sources and different data scales. Also, we might explore potential techniques to solve this issue, such as regularization~\cite{li2025preserving} and reinforcement learning~\cite{schulman2017proximal,shao2024deepseekmath,rafailov2023direct}
Second, while the PAI framework achieves strong performance on the Loong benchmark, it still relies on domain-specific human-crafted prompts to guide agents in maintaining structured reasoning. For example, prompts may explicitly direct the model to focus on concepts such as ``profit, revenue, or cash flow'' when processing financial data, or ``reference'' and ``citation'' when handling scientific texts. Therefore, we plan to explore inference methods that enable autonomous reasoning with long-context inputs, minimizing reliance on extensive human-crafted prompts.

Despite the current limitations of PAI and LongPAI, their ability to achieve significant performance gains through minimal domain-specific guidance highlights their strong practical potential for specialized applications.

% \newpage

\bibliography{custom}

\newpage

\appendix

\section{Appendix} \label{sec:appendix}

\subsection{Trend of Long-Context LLMs}
\begin{figure*}[t]
    \centering
    \includegraphics[width=1\textwidth]{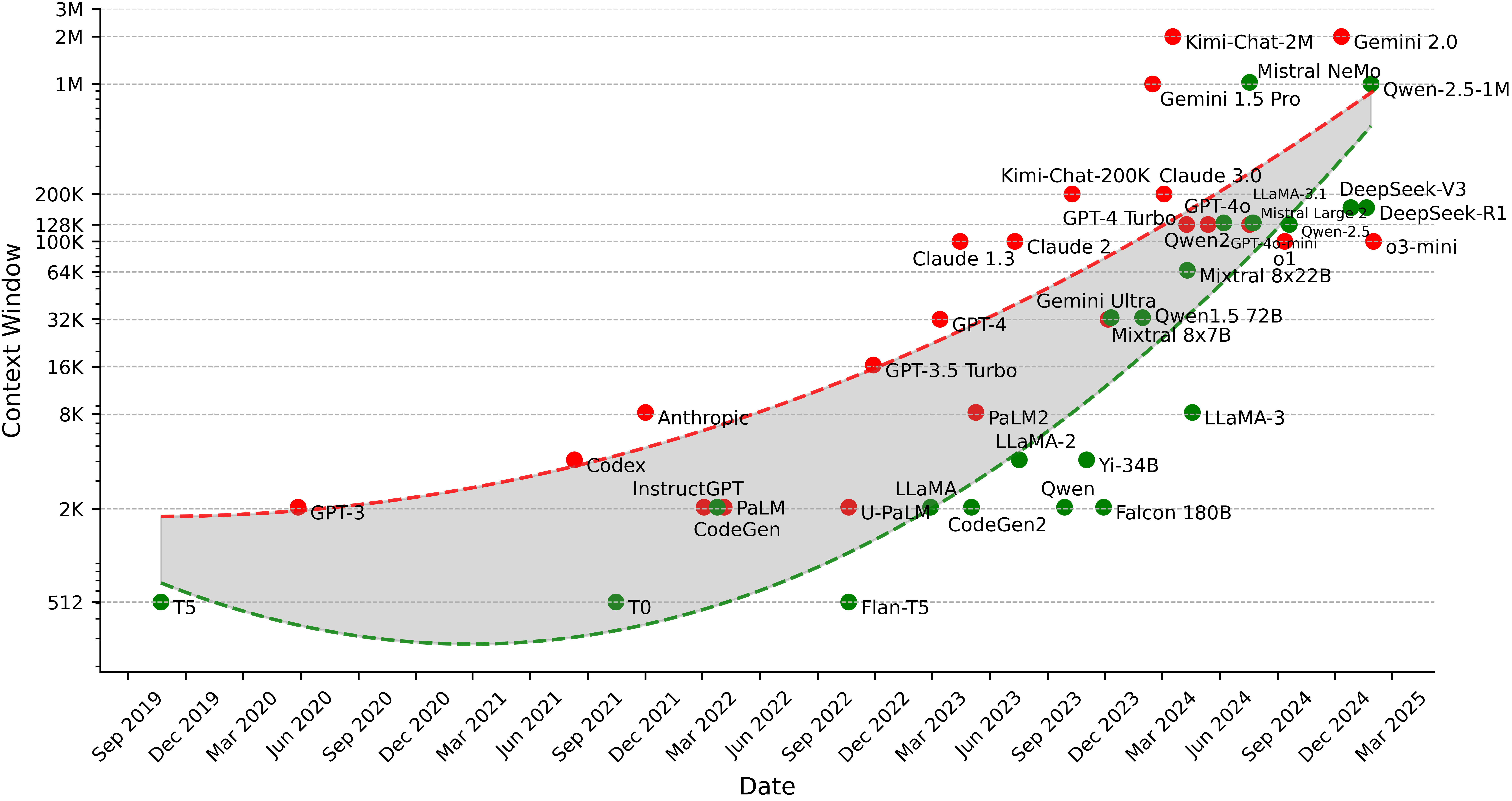}
    \caption{The trend of the context window size of recent large language models. The green dots refer to open-source LLMs, while the red dots denote closed-source LLMs.}
    \label{fig:trend}
\end{figure*}

In this section, we present the context window sizes of closed-source and open-source LLMs in Figure~\ref{fig:trend}. It indicates the gap between closed-source and open-source LLMs becomes smaller. In particular, the LLMs listed in Figure~\ref{fig:trend} include T5~\cite{raffel2020exploring}, GPT-3~\cite{brown2020language}, Codex~\cite{chen2021evaluating}, T0~\cite{sanh2022multitask}, Anthropic~\cite{priyanshu2024ai}, InstructGPT~\cite{ouyang2022training}, CodeGen~\cite{nijkamp2023codegen}, PaLM~\cite{chowdhery2023palm}, U-PaLM~\cite{tay2022transcending}, Flan-T5~\cite{chung2024scaling}, GPT-3.5 Turbo, LLaMA~\cite{touvron2023llama}, GPT-4~\cite{achiam2023gpt}, Claude 1.3, CodeGen2~\cite{nijkamp2023codegen2}, PaLM2~\cite{anil2023palm}, Claude 2, LLaMA-2~\cite{touvron2023llama}, Qwen~\cite{bai2023qwen}, Kimi-Chat-200K, Yi-34B~\cite{young2024yi}, Falcon 180B~\cite{almazrouei2023falcon}, Gemini Ultra~\cite{team2023gemini}, Mixtral~\cite{jiang2024mixtral}, Qwen1.5~\cite{qwen2024qwen1-5}, Gemini 1.5 Pro, Claude 3.0~\cite{anthropic2024claude3}, Kimi-Chat-2M, GPT-4 Turbo~\cite{achiam2023gpt}, LLaMA-3~\cite{dubey2024llama}, GPT-4o~\cite{hurst2024gpt}, Qwen2~\cite{yang2024qwen2}, GPT-4o-mini~\cite{hurst2024gpt}, Mistral NeMo~\cite{mistral2024mistralnemo}, LLaMA-3.1~\cite{dubey2024llama}, Mistral Large 2~\cite{mistral2024mistrallarge}, o1, Qwen-2.5~\cite{yang2024qwen2-5}, Gemini 2.0~\cite{pichai2024gemini2}, DeepSeek-V3~\cite{liu2024deepseek}, DeepSeek-R1~\cite{guo2025deepseek}, Qwen-2.5-1M~\cite{yang2024qwen2-5}, and o3-mini.

\subsection{Prompts of Property-based Agentic Inference (PAI)}
In this section, we present the prompts for the three agents in the PAI framework: the property extraction agent, the property-based retrieval agent, and the summarization agent.

\begin{figure}
    \centering
    \includegraphics[width=0.44\textwidth]{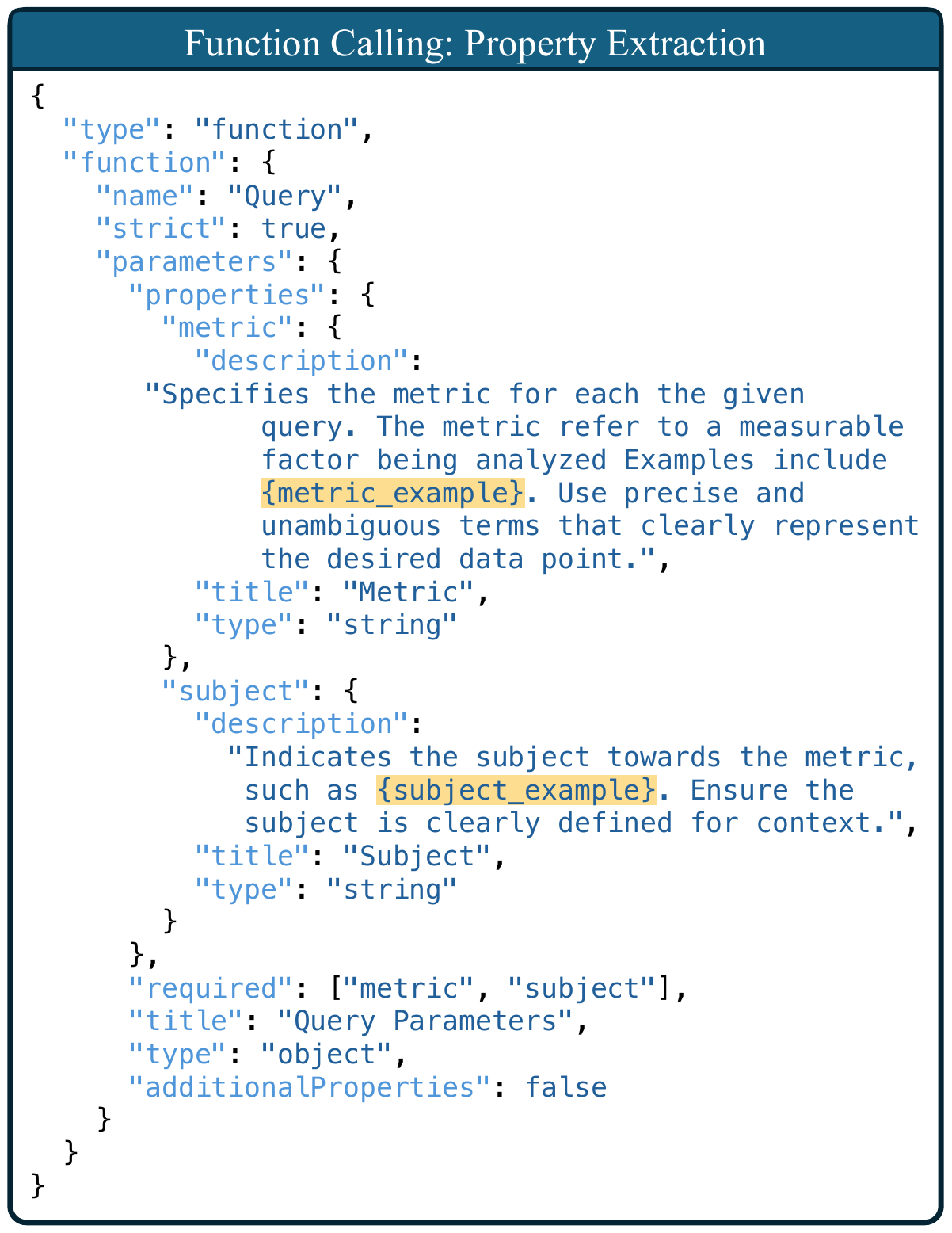}
    \includegraphics[width=0.44\textwidth]{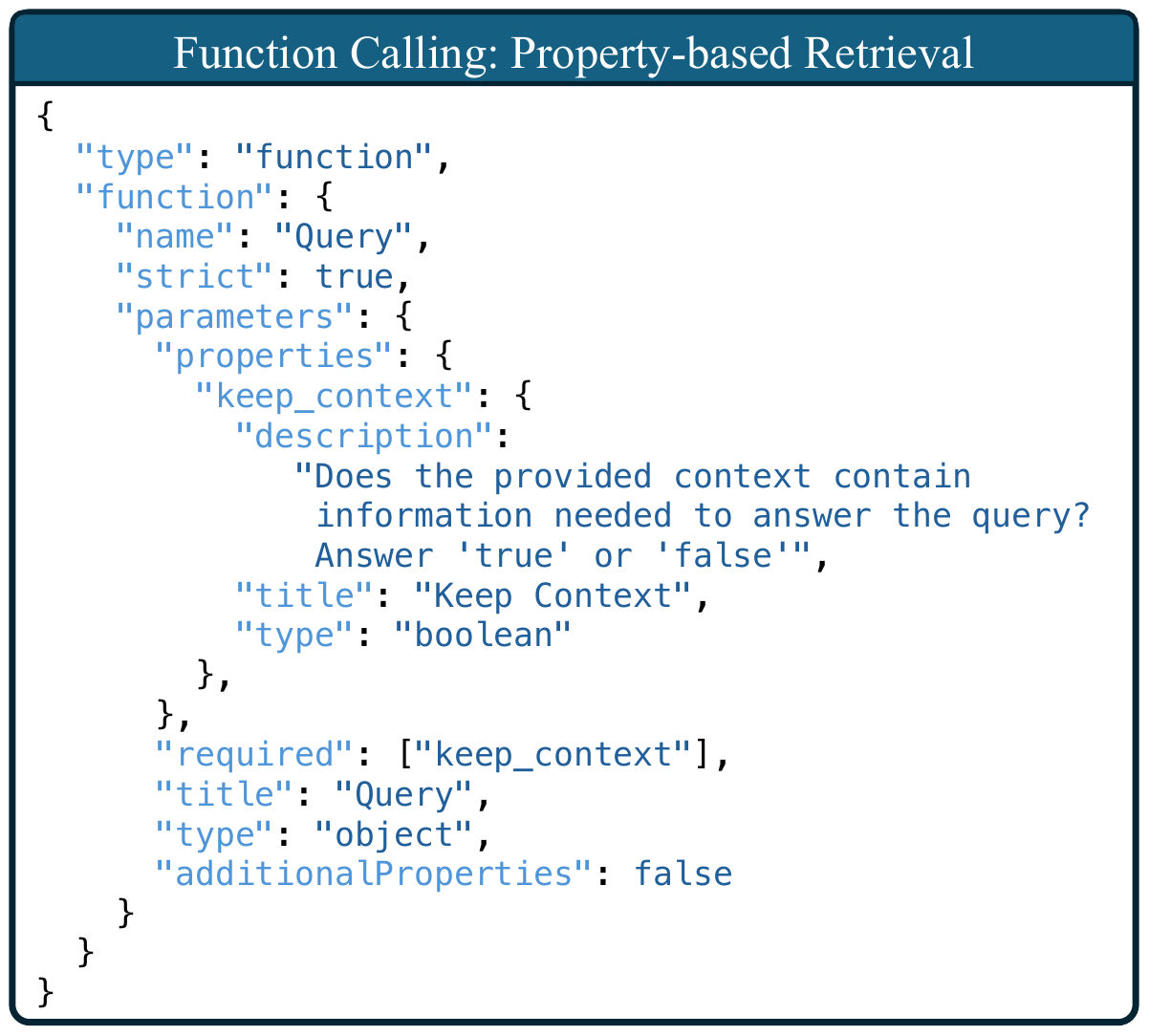}
    \caption{Function calling details of property extraction agent (\textit{top}) and property-based retrieval (\textit{bottom}).} \label{fig:pai_fc}
\end{figure}

\noindent \textbf{Property Extraction Agent}.
To extract properties, we employ the function-calling API of GPT-4o-mini to selectively retrieve the relevant metric and its corresponding subject from a given query. Figure~\ref{fig:pai_fc} illustrates the property extraction process using function calling.
In addition, we incorporate domain-specific examples within the function call to enhance accuracy across different domains. For instance, in the finance domain, we use ``profit'', ``revenue'', and ``debt'' as metric examples, while the ``financial document title'' serves as a subject example. In the legal domain, ``verdict'' represents the metric, and ``legal judgment'' serves as the subject. Similarly, in the academic domain, ``reference'' and ``citation'' are used as metric examples, with ``paper title'' as the subject.

\begin{figure}
    \centering
    \includegraphics[width=0.2\textwidth]{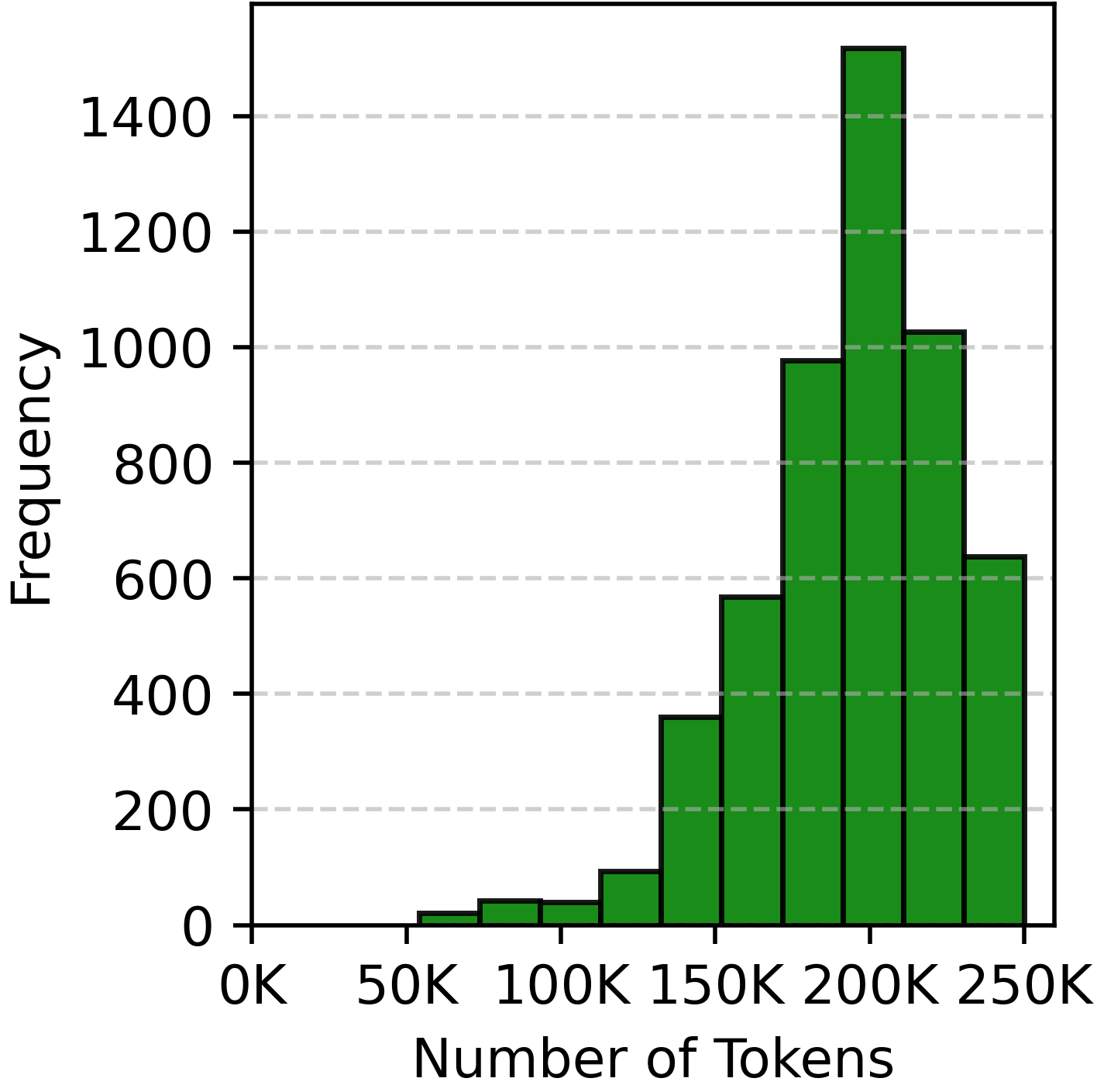}
    \includegraphics[width=0.27\textwidth]{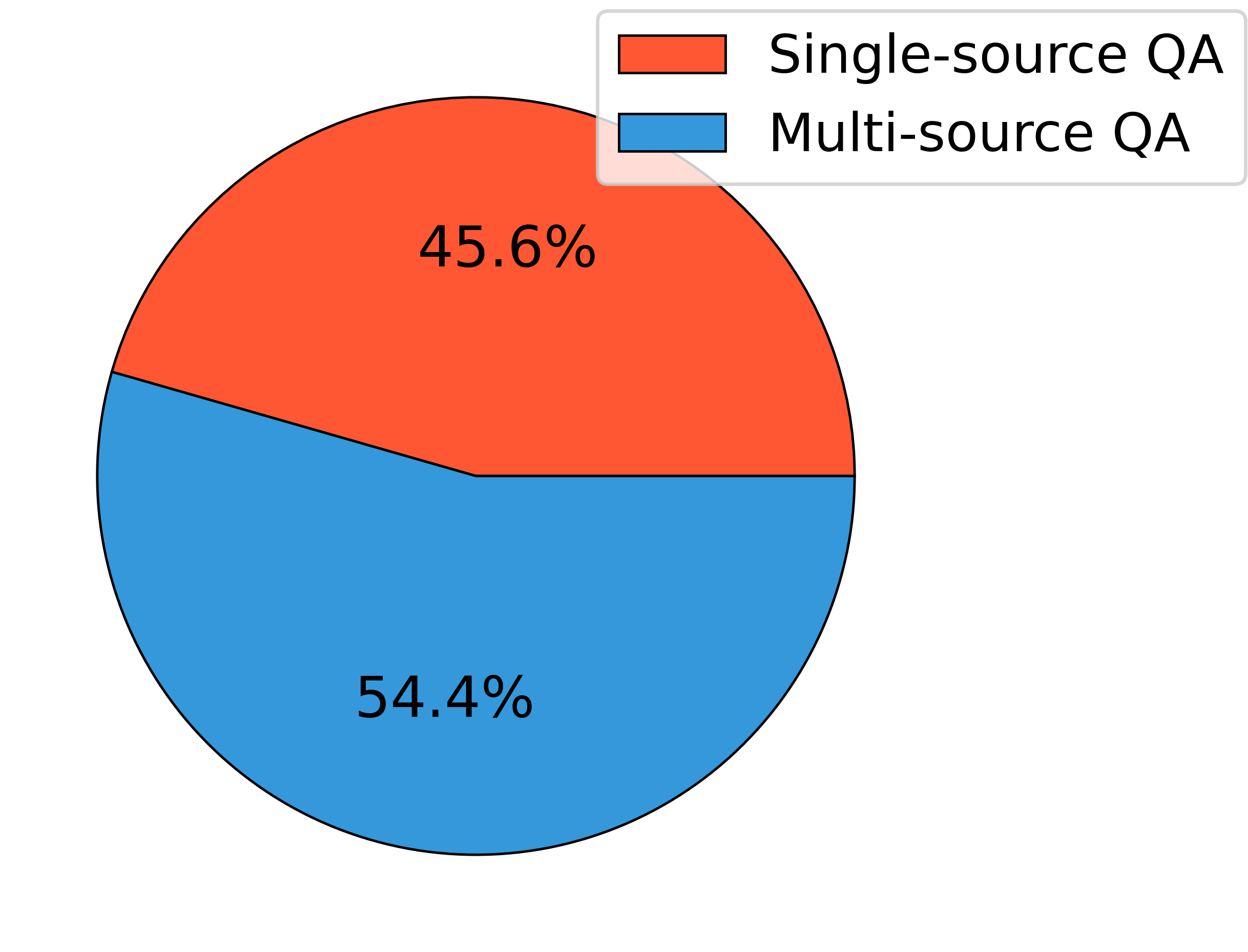}
    \caption{The histogram of input token length in the LongFinanceQA (\textit{left}). The proportion of single-source and multi-source QA tasks in the LongFinanceQA (\textit{right}).}
    \label{fig:data_stat}
\end{figure}

\begin{figure}
    \centering
    \includegraphics[width=0.4\textwidth]{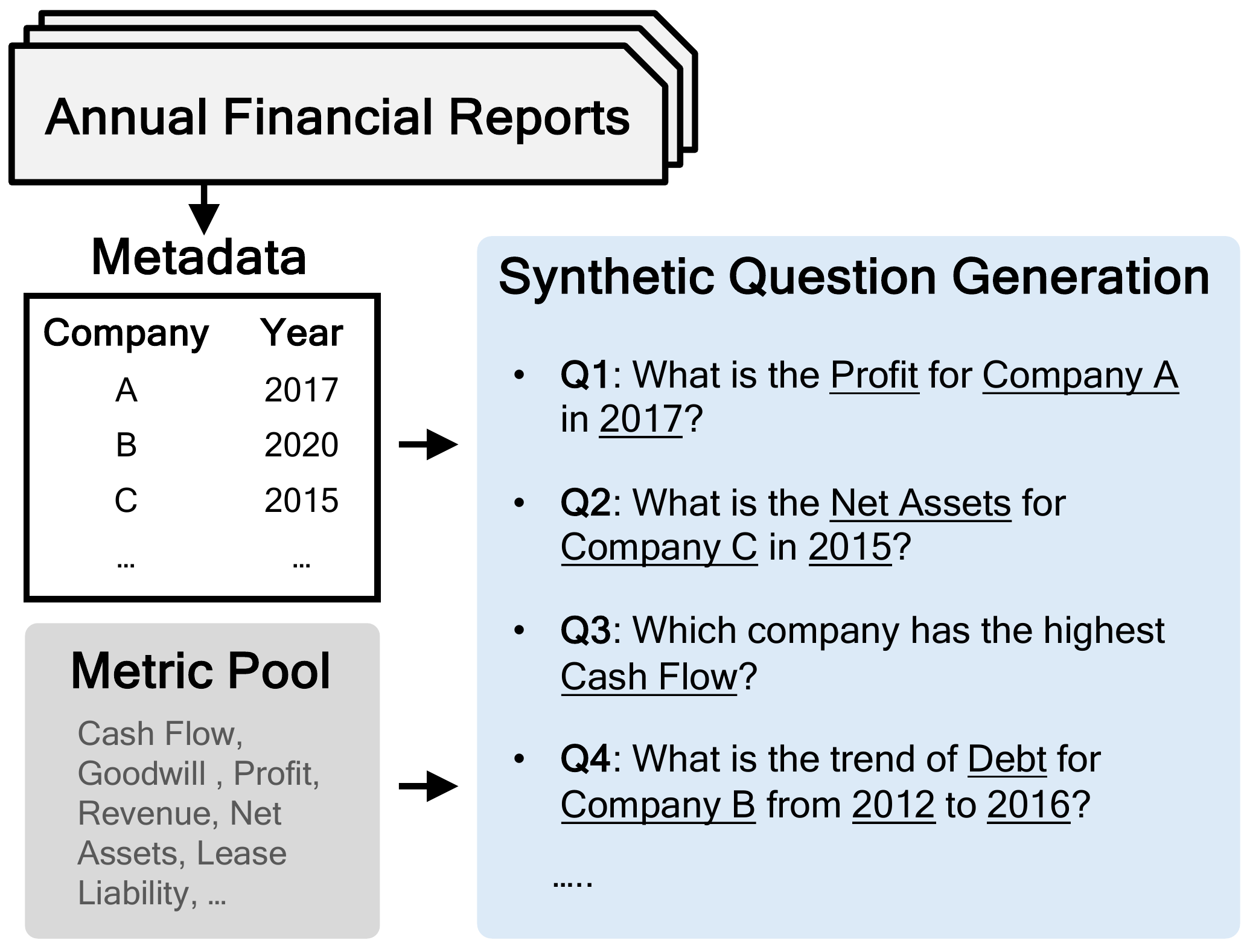}
    \caption{The pipeline of long-context synthetic question generation for the LongFinanaceQA dataset.}
    \label{fig:query_gen}
\end{figure}

\noindent \textbf{Property-based Retrieval Agent}. After obtaining the metric and its corresponding subject, we generate a sub-question in the format: ``What was the <metric> of the <subject>?''. Then, the long-context input document is divided into a list of 1024-token chunks. Each chunk is evaluated to determine its relevance to the sub-questions, as shown in Figure~\ref{fig:pai_fc} (\textit{bottom}). After that, each sub-question is assigned relevant chunks. Next, we pack these relevant chunks and generate a sub-answer to the corresponding sub-question. Thus, this agent functions similarly to RAG~\cite{lewis2020retrieval}.

\noindent \textbf{Summarization Agent}. Given the original query, the summarization agent summarizes a conclusion based on the sub-answers generated by the property-based retrieval agent.

\subsection{LongFinanceQA Dataset Statistics}
Figure~\ref{fig:data_stat} presents a histogram of input token lengths in the LongFinanceQA dataset, which generally range from 50K to 250K tokens. Most input documents contain approximately 200K tokens. At the same time, we provide the proportion of single-source and multi-source QA pairs in the proposed LongFinanceQA dataset, where single-source QA pairs make up 45.6\%, while multi-source QA pairs account for 54.4\%.

\begin{figure*}[t]
    \centering
    \includegraphics[width=0.95\textwidth]{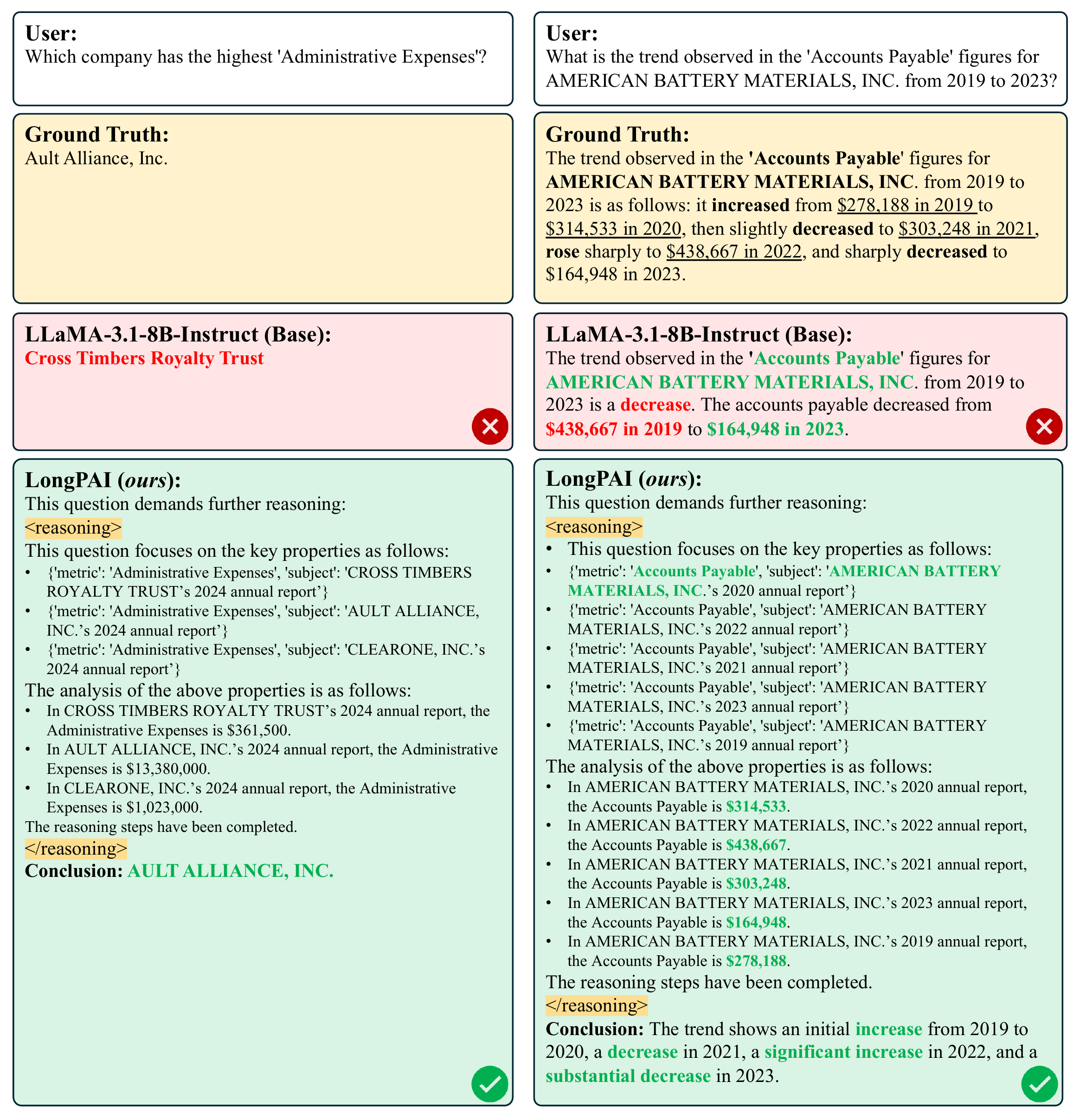}
    \caption{Case study on two representative long-context QA problems.}
    \label{fig:case_study}
\end{figure*}

\subsection{Synthetic Question Generation}
Figure~\ref{fig:query_gen} illustrates the procedure of synthetic Question generation. First, We create a financial metric pool with key metrics like profit, revenue, and cash flow (represented as the grey box). Then, we randomly select metrics and metadata of financial reports (\eg, company name and year) to generate long-context questions requiring single- or multi-source evidence (represented as the blue box). In addition, the types of multi-source questions are inspired by Loong~\cite{wang2024leave} and include comparison, clustering, trend analysis, etc.

\subsection{More Empirical Results}

\noindent \textbf{Case Study on Practical Long-Context Problems}.
As shown in Figure~\ref{fig:case_study}, we present two representative long-context QA examples. Both questions are particularly challenging, as they require models to consider multiple pieces of evidence from the input text. Specifically, the case study includes the user query (white box), ground truth (yellow box), predictions from the base LLaMA-3.1-8B-Instruct model~\cite{dubey2024llama} (white box), and predictions from our LongPAI model (green box). In addition, incorrect predictions are highlighted in red, while correct ones are marked in green. The results demonstrate the superiority of the proposed LongPAI model, which benefits from supervised CoT reasoning. Furthermore, Figure~\ref{fig:case_study} illustrates that LongPAI offers significantly greater interpretability than its base model.

% \noindent \textbf{The impact of $K$ in RAG}.
% We present in Figure~, 

% 

% metric pool

\subsection{Full Performances}
Table~\ref{tab:full_longpai_results} presents the full results of the Table~\ref{tab:longpai_results} in the main manuscript. Also, we involve more popular LLMs~\cite{yang2024qwen2-5,guo2025deepseek} into the Table~\ref{tab:full_longpai_results}.

\begin{table*}[t]
\small
\centering  
\caption{Full results of Table~\ref{tab:longpai_results}. \textit{AS} represents \textit{Avg Scores (0\textasciitilde100)} and \textit{{PR}} denotes \textit{Perfect Rate} (0\textasciitilde1). \colorbox{mygreen}{Green} indicates improvements compared to the base model; \colorbox{myred}{Red} denotes a decrease compared to the base model.}
\renewcommand{\arraystretch}{1}
\resizebox{\textwidth}{!}{
\begin{tabular}{llcccccccccc}
\toprule
\multirow{2}{*}{\textbf{Model}} & \multicolumn{1}{c}{\textbf{Context}} & \multicolumn{2}{c}{\textbf{Spotlight Locating}} & \multicolumn{2}{c}{\textbf{Comparison}} & \multicolumn{2}{c}{\textbf{Clustering}} & \multicolumn{2}{c}{\textbf{Chain of Reasoning}} & \multicolumn{2}{c}{\textbf{Overall}}\\ \cmidrule(r){3-4} \cmidrule(r){5-6} \cmidrule(r){7-8} \cmidrule(r){9-10} \cmidrule(r){11-12}
 & \multicolumn{1}{c}{\textbf{Length}} & \textbf{\textit{AS}} & \textbf{\textit{PR}} & \textbf{\textit{AS}} & \textbf{\textit{PR}} & \textbf{\textit{AS}} & \textbf{\textit{PR}} & \textbf{\textit{AS}} & \textbf{\textit{PR}} & \textbf{\textit{AS}} & \textbf{\textit{PR}} \\

\midrule
\multicolumn{12}{>{\columncolor[gray]{.88}}c}{\textbf{$\mathtt{All\ Set}$ (10K-250K)}}  \\
DeepSeek-R1-Qwen-32B & 128K &  53.66 & 0.46 & 52.19 & 0.39 & 39.76 & 0.17 & 65.15 & 0.51 & 49.92 & 0.34 \\
Qwen2-72B-Instruct & 128K &  59.80 & 0.47 & 61.12 & 0.43 & 34.32 & 0.06 & 74.68 & 0.50 & 53.20 & 0.32  \\
Qwen2.5-72B-Instruct & 128K &  71.07 & 0.63 & 59.14 & 0.41 & 38.23 & 0.08 & 81.09 & 0.60 & 57.36 & 0.36  \\
LLaMA-3-8B-Instruct-262K & 262K  &  58.60 & 0.41 & 33.12 & 0.16 & 20.04 & 0.01 & 35.10 & 0.09 & 34.41 & 0.15 \\
GLM4-9B-Chat & 1000K & 72.69 & 0.60 & 49.31 & 0.32 & 23.41 & 0.02 & 60.77 & 0.28 & 46.71 & 0.27 \\
Qwen-2.5-14B-Instruct-1M & 1000K & 78.83 & 0.72 & 65.27 & 0.50 & 36.24 & 0.07 & 79.10 & 0.64 & 59.78 & 0.41 \\
GPT-4o & 128K &  88.23 & 0.84 & 62.90 & 0.48 & 45.51 & 0.17 & 69.40 & 0.43 & 63.05 & 0.44 \\
GPT-4o-mini (Base) & 128K & 70.90 & 0.59 & 59.37 & 0.38 & 36.33 & 0.06 & 79.58 & 0.58 & 56.50 & 0.34 \\
GPT-4o-mini w/ PAI (\textit{ours}) & 128K & \cellcolor{mygreen}91.07 & \cellcolor{mygreen}0.83 & \cellcolor{mygreen}74.40 & \cellcolor{mygreen}0.58 & \cellcolor{mygreen}61.55 & \cellcolor{mygreen}0.32 & \cellcolor{mygreen}89.63 & \cellcolor{mygreen}0.78 & \cellcolor{mygreen}75.56 & \cellcolor{mygreen}0.57 \\
\midrule
LLaMA-3.1-8B-Instruct (Base) & 128K & 67.84 & 0.56 & 47.12 & 0.30 & 24.62 & 0.02 & 63.63 & 0.34 & 45.88 & 0.26 \\ 
LongPAI (\textit{ours}) & 262K & \cellcolor{mygreen}89.79 & \cellcolor{mygreen}0.84 & \cellcolor{mygreen}71.69 & \cellcolor{mygreen}0.60 & \cellcolor{mygreen}59.71 & \cellcolor{mygreen}0.32 & \cellcolor{mygreen}90.28 & \cellcolor{mygreen}0.83 & \cellcolor{mygreen}73.94 & \cellcolor{mygreen}0.58 \\

\midrule
\multicolumn{12}{>{\columncolor[gray]{.88}}c}{\textbf{$\mathtt{Set1}$ (10K-50K)}}  \\
DeepSeek-R1-Qwen-32B & 128K &  41.73 & 0.33 & 39.56 & 0.23 & 25.67 & 0.06 & 55.14 & 0.37 & 37.35 & 0.21 \\
Qwen2-72B-Instruct & 128K &  88.04 & 0.83 & 89.33 & 0.83 & 43.00 & 0.17 & 93.50 & 0.80 & 71.46 & 0.57  \\
Qwen2.5-72B-Instruct & 128K &  88.70 & 0.87 & 84.67 & 0.80 & 43.92 & 0.10 & 88.00 & 0.80 & 70.07 & 0.54  \\
LLaMA-3-8B-Instruct-262K & 262K  &  75.82 & 0.64 & 40.83 & 0.23 & 20.68 & 0.03 & 68.00 & 0.40 & 43.14 & 0.25 \\
GLM4-9B-Chat & 1000K & 88.26 & 0.83 & 73.17 & 0.57 & 24.65 & 0.00 & 87.30 & 0.50 & 59.07 & 0.40 \\
Qwen-2.5-14B-Instruct-1M & 1000K &  96.09 & 0.96 & 88.67 & 0.80 & 49.00 & 0.20 & 100.00 & 1.00 & 76.02 & 0.62 \\
GPT-4o & 128K &  100.00 & 1.00 & 88.50 & 0.80 & 55.25 & 0.28 & 98.50 & 0.90 & 79.13 & 0.65 \\
GPT-4o-mini (Base) & 128K & 97.39 & 0.96 & 81.83 & 0.67 & 46.50 & 0.15 & 100.00 & 1.00 & 73.35 & 0.56 \\
GPT-4o-mini w/ PAI (\textit{ours}) & 128K & \cellcolor{myred}95.87 & \cellcolor{myred}0.91 & \cellcolor{mygreen}91.50 & \cellcolor{mygreen}0.83 & \cellcolor{mygreen}78.38 & \cellcolor{mygreen}0.60 & \cellcolor{myred}96.80 & \cellcolor{myred}0.70 & \cellcolor{mygreen}87.89 & \cellcolor{mygreen}0.75 \\
\midrule
LLaMA-3.1-8B-Instruct  (Base) & 128K & 89.13 & 0.87 & 72.33 & 0.60 & 31.77 & 0.05 & 74.00 & 0.60 & 60.50 & 0.45 \\ 
LongPAI (\textit{ours}) & 262K & \cellcolor{mygreen}97.30 & \cellcolor{mygreen}0.91 & \cellcolor{mygreen}90.17 & \cellcolor{mygreen}0.80 & \cellcolor{mygreen}72.88 & \cellcolor{mygreen}0.47 & \cellcolor{mygreen}94.00 & \cellcolor{mygreen}0.90 & \cellcolor{mygreen}85.42 & \cellcolor{mygreen}0.71 \\

\midrule
\multicolumn{12}{>{\columncolor[gray]{.88}}c}{\textbf{$\mathtt{Set2}$ (50K-100K)}}  \\
DeepSeek-R1-Qwen-32B & 128K &  41.73 & 0.33 & 39.56 & 0.23 & 25.67 & 0.06 & 55.14 & 0.37 & 37.35 & 0.21 \\
Qwen2-72B-Instruct & 128K &  74.88 & 0.65 & 68.60 & 0.51 & 40.70 & 0.09 & 87.00 & 0.72 & 62.38 & 0.41  \\
Qwen2.5-72B-Instruct & 128K &  86.00 & 0.82 & 61.64 & 0.41 & 46.94 & 0.16 & 93.12 & 0.85 & 65.36 & 0.46  \\
LLaMA-3-8B-Instruct-262K & 262K  &  72.44 & 0.54 & 44.03 & 0.26 & 24.11 & 0.02 & 37.25 & 0.12 & 40.40 & 0.20 \\
GLM4-9B-Chat & 1000K & 82.12 & 0.68 & 52.73 & 0.36 & 26.04 & 0.04 & 76.83 & 0.42 & 51.66 & 0.31 \\
Qwen-2.5-14B-Instruct-1M & 1000K &  88.75 & 0.85 & 74.07 & 0.61 & 39.06 & 0.09 & 96.38 & 0.90 & 67.24 & 0.51 \\
GPT-4o & 128K &  95.75 & 0.95 & 72.87 & 0.57 & 54.94 & 0.27 & 73.38 & 0.47 & 70.10 & 0.51 \\
GPT-4o-mini  (Base) & 128K & 82.65 & 0.70 & 58.77 & 0.41 & 40.89 & 0.08 & 92.50 & 0.78 & 61.61 & 0.40 \\
GPT-4o-mini w/ PAI (\textit{ours}) & 128K & \cellcolor{mygreen}89.38 & \cellcolor{mygreen}0.85 & \cellcolor{mygreen}70.73 & \cellcolor{mygreen}0.51 & \cellcolor{mygreen}64.37 & \cellcolor{mygreen}0.36 & \cellcolor{mygreen}94.62 & \cellcolor{mygreen}0.90 & \cellcolor{mygreen}75.34 & \cellcolor{mygreen}0.57 \\
\midrule
LLaMA-3.1-8B-Instruct (Base) & 128K & 80.58 & 0.70 & 51.11 & 0.33 & 27.39 & 0.02 & 75.12 & 0.47 & 51.13 & 0.30 \\ 
LongPAI (\textit{ours}) & 262K & \cellcolor{mygreen}91.25  & \cellcolor{mygreen}0.88 & \cellcolor{mygreen}76.27 & \cellcolor{mygreen}0.67 & \cellcolor{mygreen}66.02 & \cellcolor{mygreen}0.37 & \cellcolor{mygreen}96.12 & \cellcolor{mygreen}0.93 & \cellcolor{mygreen}78.19 & \cellcolor{mygreen}0.63\\ 

\midrule
\multicolumn{12}{>{\columncolor[gray]{.88}}c}{\textbf{$\mathtt{Set3}$ (100K-200K)}}  \\
DeepSeek-R1-Qwen-32B & 128K &  41.73 & 0.33 & 39.56 & 0.23 & 25.67 & 0.06 & 55.14 & 0.37 & 37.35 & 0.21 \\
Qwen2-72B-Instruct & 128K &  47.00 & 0.33 & 48.07 & 0.27 & 25.79 & 0.00 & 69.37 & 0.34 & 42.98 & 0.20  \\
Qwen2.5-72B-Instruct & 128K &  60.47 & 0.48 & 49.00 & 0.28 & 30.61 & 0.01 & 76.54 & 0.46 & 48.99 & 0.26  \\
LLaMA-3-8B-Instruct-262K & 262K  &  54.14 & 0.40 & 22.93 & 0.05 & 15.43 & 0.00 & 29.00 & 0.00 & 28.58 & 0.11 \\
GLM4-9B-Chat & 1000K & 74.75 & 0.65 & 41.63 & 0.24 & 21.99 & 0.01 & 49.86 & 0.17 & 43.58 & 0.25 \\
Qwen-2.5-14B-Instruct-1M & 1000K &  74.33 & 0.68 & 54.64 & 0.35 & 30.72 & 0.01 & 73.71 & 0.51 & 53.47 & 0.33 \\
GPT-4o & 128K &  87.25 & 0.83 & 46.00 & 0.31 & 36.68 & 0.08 & 64.57 & 0.40 & 54.79 & 0.36 \\
GPT-4o-mini (Base) & 128K & 63.05 & 0.53 & 53.48 & 0.24 & 29.80 & 0.01 & 72.37 & 0.46 & 50.03 & 0.26 \\
GPT-4o-mini w/ PAI (\textit{ours}) & 128K & \cellcolor{mygreen}94.08 & \cellcolor{mygreen}0.83 & \cellcolor{mygreen}74.13 & \cellcolor{mygreen}0.63 & \cellcolor{mygreen}55.78 & \cellcolor{mygreen}0.20 & \cellcolor{mygreen}87.71 & \cellcolor{mygreen}0.77 & \cellcolor{mygreen}74.21 & \cellcolor{mygreen}0.55\\
\midrule
LLaMA-3.1-8B-Instruct (Base) & 128K & 63.38 & 0.47 & 36.04 & 0.19 & 20.28 & 0.00 & 62.49 & 0.26 & 40.45 & 0.20 \\ 
LongPAI (\textit{ours}) & 262K & \cellcolor{mygreen}94.17 & \cellcolor{mygreen}0.90 & \cellcolor{mygreen}63.76 & \cellcolor{mygreen}0.49 & \cellcolor{mygreen}51.83 & \cellcolor{mygreen}0.24 & \cellcolor{mygreen}88.66 & \cellcolor{mygreen}0.77 & \cellcolor{mygreen}70.00 & \cellcolor{mygreen}0.54 \\

\midrule
\multicolumn{12}{>{\columncolor[gray]{.88}}c}{\textbf{$\mathtt{Set4}$ (200K-250K)}}  \\
DeepSeek-R1-Qwen-32B & 128K &  18.33 & 0.11 & 10.25 & 0.05 & 11.27 & 0.00 & 26.00 & 0.07 & 15.52 & 0.05 \\
Qwen2-72B-Instruct & 128K &  41.85 & 0.19 & 39.75 & 0.15 & 29.17 & 0.03 & 41.67 & 0.07 & 37.23 & 0.11  \\
Qwen2.5-72B-Instruct & 128K &  57.48 & 0.44 & 49.50 & 0.30 & 27.33 & 0.00 & 55.00 & 0.13 & 45.51 & 0.22  \\
LLaMA-3-8B-Instruct-262K & 262K  &  34.19 & 0.07 & 20.00 & 0.05 & 20.31 & 0.00 & 20.71 & 0.00 & 24.61 & 0.03 \\
GLM4-9B-Chat & 1000K & 40.85 & 0.19 & 29.50 & 0.05 & 18.13 & 0.00 & 25.73 & 0.00 & 28.51 & 0.07 \\
Qwen-2.5-14B-Instruct-1M & 1000K &  59.44 & 0.41 & 37.00 & 0.20 & 27.33 & 0.00 & 31.67 & 0.00 & 39.57 & 0.16 \\
GPT-4o & 128K &  69.26 & 0.56 & 50.50 & 0.35 & 30.70 & 0.00 & 50.67 & 0.07 & 49.58 & 0.25 \\
GPT-4o-mini (Base) & 128K & 48.37 & 0.26 & 50.00 & 0.30 & 28.70 & 0.00 & 48.33 & 0.07 & 42.30 & 0.15 \\
GPT-4o-mini w/ PAI (\textit{ours}) & 128K & \cellcolor{mygreen}82.78 & \cellcolor{mygreen}0.70 & \cellcolor{mygreen}63.50 & \cellcolor{mygreen}0.35 & \cellcolor{mygreen}48.00 & \cellcolor{mygreen}0.17 & \cellcolor{mygreen}76.00 & \cellcolor{mygreen}0.53 & \cellcolor{mygreen}66.14 & \cellcolor{mygreen}0.42\\
\midrule
LLaMA-3.1-8B-Instruct (Base) & 128K & 40.74 & 0.30 & 35.85 & 0.20 & 19.77 & 0.00 & 28.73 & 0.00 & 30.88 & 0.13 \\ 
LongPAI (\textit{ours})  & 262K & \cellcolor{mygreen}71.48 & \cellcolor{mygreen}0.59 & \cellcolor{mygreen}56.50 & \cellcolor{mygreen}0.45 & \cellcolor{mygreen}46.83 & \cellcolor{mygreen}0.17 & \cellcolor{mygreen}76.00 & \cellcolor{mygreen}0.67 & \cellcolor{mygreen}60.92 & \cellcolor{mygreen}0.43 \\ 

\bottomrule
\end{tabular}\label{tab:full_longpai_results}
}
\end{table*}

\end{document}